\documentclass[10pt,twocolumn,letterpaper]{article}

\usepackage[pagenumbers]{cvpr} 

\usepackage{graphicx}
\usepackage{amsmath}
\usepackage{amssymb}

\usepackage{algorithm}
\usepackage{algorithmic}
\usepackage{bm}
\usepackage{mathtools}
\usepackage{makecell}
\usepackage{enumitem}
\usepackage{caption}
\usepackage{lipsum}
\usepackage{cuted}
\usepackage{siunitx}
\usepackage{booktabs}

\usepackage{subcaption}
\usepackage{multirow}

\newcommand{\vect}[1]{\boldsymbol{\mathbf{#1}}}

\usepackage[pagebackref,breaklinks,colorlinks]{hyperref}

\usepackage[capitalize]{cleveref}
\crefname{section}{Sec.}{Secs.}
\Crefname{section}{Section}{Sections}
\Crefname{table}{Table}{Tables}
\crefname{table}{Tab.}{Tabs.}

\def\cvprPaperID{07053} 
\def\confName{CVPR}
\def\confYear{2022}

\begin{document}
	
	
	\title{Scaling Up Your Kernels to 31x31: Revisiting Large Kernel Design in CNNs}
	
	\author{Xiaohan Ding \textsuperscript{1}\thanks{This work is supported by the National Natural Science
			Foundation of China (Nos.61925107, U1936202, 62021002) and the Beijing Academy of Artificial Intelligence (BAAI). This work is done during Xiaohan Ding's internship at MEGVII Technology.} 
		\quad Xiangyu Zhang \textsuperscript{2}\thanks{Project leader and corresponding author.}  
		\quad Yizhuang Zhou \textsuperscript{2} \\
		Jungong Han \textsuperscript{3} 
		\quad Guiguang Ding \textsuperscript{1}
		\quad Jian Sun \textsuperscript{2} \\
		\textsuperscript{1} Beijing National Research Center for Information Science and Technology (BNRist); \\School of Software, Tsinghua University, Beijing, China \\
		\textsuperscript{2} MEGVII Technology \\
		\textsuperscript{3} Computer Science Department, Aberystwyth University, SY23 3FL, UK \\
		\tt\small dxh17@mails.tsinghua.edu.cn \quad zhangxiangyu@megvii.com \quad zhouyizhuang@megvii.com\\
		\tt\small jungonghan77@gmail.com \quad dinggg@tsinghua.edu.cn \quad sunjian@megvii.com \\
	}
	
	\maketitle

	\begin{abstract}
	    We revisit large kernel design in modern convolutional neural networks (CNNs). Inspired by recent advances in vision transformers (ViTs), in this paper, we demonstrate that using a few large convolutional kernels instead of a stack of small kernels could be a more powerful paradigm. We suggested five guidelines, \eg, applying re-parameterized large depth-wise convolutions, to design efficient high-performance large-kernel CNNs. Following the guidelines, we propose RepLKNet, a \textbf{pure} CNN architecture whose kernel size is as large as 31$\times$31, in contrast to commonly used 3$\times$3. RepLKNet greatly closes the performance gap between CNNs and ViTs, \eg, achieving comparable or superior results than Swin Transformer on ImageNet and a few typical downstream tasks, with lower latency. RepLKNet also shows nice scalability to big data and large models, obtaining \textbf{87.8\%} top-1 accuracy on ImageNet and \textbf{56.0\%} mIoU on ADE20K, which is very competitive among the state-of-the-arts with similar model sizes. Our study further reveals that, in contrast to small-kernel CNNs, large-kernel CNNs have much larger effective receptive fields and higher shape bias rather than texture bias. Code \& models at \url{https://github.com/megvii-research/RepLKNet}.
	\end{abstract}
	
	\section{Introduction}
	\label{sec:intro}

	\begin{figure}[t]
		\begin{center}
			\includegraphics[width=\linewidth]{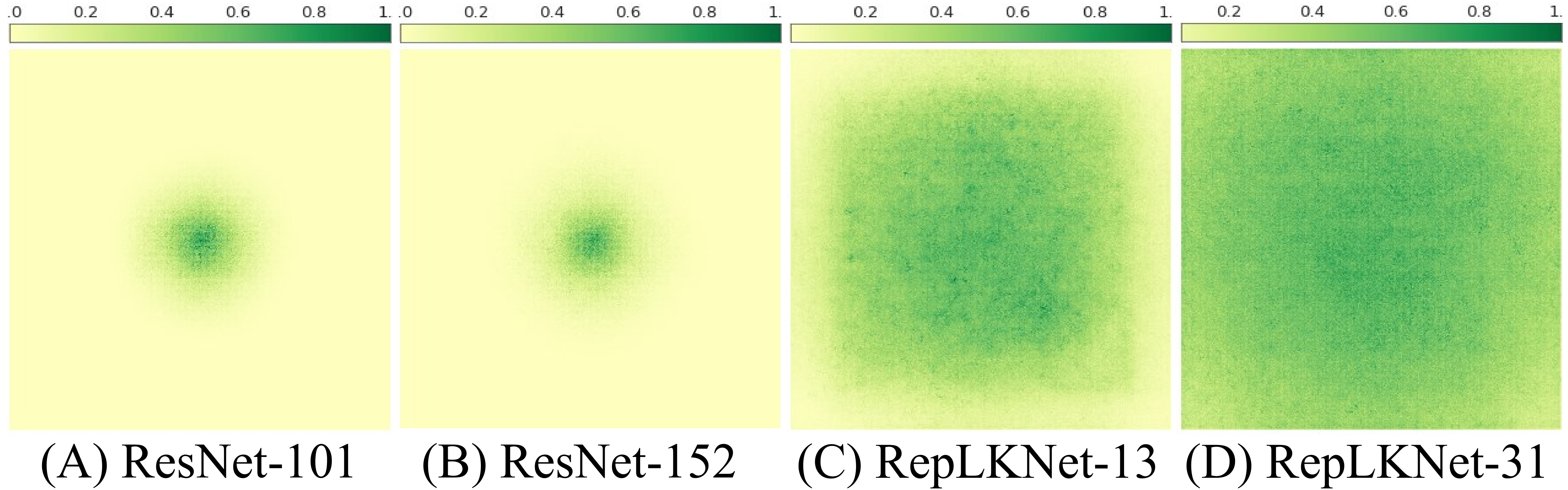}
			\vspace{-0.25in}
			\caption{The \emph{Effective Receptive Field (ERF)} of ResNet-101/152 and RepLKNet-13/31 respectively. A more widely distributed dark area indicates a larger ERF. More layers (\eg, from ResNet-101 to ResNet-152) help little in enlarging ERFs. Instead, our large kernel model \emph{RepLKNet} effectively obtains large ERFs.}
			\label{fig-rf}
			\vspace{-0.35in}
		\end{center}
	\end{figure}	
	
	\emph{Convolutional neural networks (CNNs)} \cite{krizhevsky2012imagenet,he2016deep} used to be a common choice of visual encoders in modern computer vision systems. However, recently, CNNs \cite{krizhevsky2012imagenet,he2016deep} have been greatly challenged by \emph{Vision Transformers (ViTs)}~\cite{vit,swin,deit,pvt}, which have shown leading performances on many visual tasks -- not only image classification~\cite{vit,yuan2021volo} and representation learning~\cite{mocov3,dino,swinself,bao2021beit}, but also many downstream tasks such as object detection~\cite{swin,dai2021dynamic}, semantic segmentation~\cite{pvt,xie2021segformer} and image restoration~\cite{ipt,liang2021swinir}. 
	Why are ViTs super powerful? Some works believed that \emph{multi-head self-attention (MHSA)} mechanism in ViTs plays a key role. They provided empirical results to demonstrate that, MHSA is more flexible~\cite{clip}, capable (less inductive bias)~\cite{cordonnier2019relationship}, more robust to distortions~\cite{paul2021vision,xie2021segformer}, or able to model long-range dependencies~\cite{vaswani2017attention,raghu2021vision}. But some works challenge the necessity of MHSA \cite{zhu2019empirical}, attributing the high performance of ViTs to the proper building blocks~\cite{dong2021attention}, and/or dynamic sparse weights~\cite{han2021demystifying,zhao2021battle}. More works~\cite{hinton2021represent,zhu2019empirical,han2021demystifying,wu2019pay,cordonnier2019relationship} explained the superiority of ViTs from different point of views.
	
	In this work, we focus on one view: \emph{the way of building up large receptive fields}. In ViTs, \emph{MHSA} is usually designed to be either global~\cite{vit,pvt,bot} or local but with large kernels~\cite{swin,halonet,sasa}, thus each output from a \emph{single} MHSA layer is able to gather information from a large region. However, large kernels are not popularly employed in CNNs (except for the first layer~\cite{he2016deep}). Instead, a typical fashion is to use a stack of many small \emph{spatial convolutions}\footnote{Convolutional kernels (including the variants such as depth-wise/group convolutions) whose spatial size is larger than 1$\times$1.} \cite{simonyan2014very,he2016deep,zhang2018shufflenet,mbv1,huang2017densely,efficientnet,regnet} (\eg, 3$\times$3) to enlarge the receptive fields in state-of-the-art CNNs. Only some old-fashioned networks such as \emph{AlexNet}~\cite{krizhevsky2012imagenet}, \emph{Inceptions}~\cite{szegedy2015going,szegedy2016rethinking,szegedy2017inception} and a few architectures derived from \emph{neural architecture search}~\cite{mbv3,liu2018darts,zoph2016neural,guo2020single} adopt large spatial convolutions (whose size is greater than 5) as the main part. 
	The above view naturally lead to a question: what if we use \emph{a few large} instead of \emph{many small} kernels to conventional CNNs? Is large kernel or the way of building large receptive fields the key to close the performance gap between CNNs and ViTs?

	To answer this question, we systematically explore the large kernel design of CNNs. We follow a very simple ``philosophy'': just introducing large \emph{depth-wise} convolutions into conventional networks, whose sizes range from 3$\times$3 to 31$\times$31, although there exist other alternatives to introduce large receptive fields via a single or a few layers, \eg feature pyramids~\cite{wang2020deep}, dilated convolutions~\cite{yu2015multi,yu2017dilated,chen2017deeplab} and deformable convolutions~\cite{dai2017deformable}. Through a series of experiments, we summarize five empirical guidelines to effectively employ large convolutions: \textbf{1)} very large kernels can still be efficient in practice; \textbf{2)} identity shortcut is vital especially for networks with very large kernels; \textbf{3)} re-parameterizing~\cite{ding2021repvgg} with small kernels helps to make up the optimization issue; \textbf{4)} large convolutions boost downstream tasks much more than \emph{ImageNet}; \textbf{5)} large kernel is useful even on small feature maps.

	Based on the above guidelines, we propose a new architecture named \emph{RepLKNet}, a \emph{pure}\footnote{Namely CNNs free of any attention or dynamic mechanism, \eg, \emph{squeeze-and-excitation}~\cite{hu2018squeeze}, \emph{multi-head self-attention}, \emph{dynamic weights}~\cite{han2021demystifying,wu2019pay}, and \etc.} CNN where re-parameterized large convolutions are employed to build up large receptive fields. Our network in general follows the macro architecture of \emph{Swin Transformer}~\cite{swin} with a few modifications, while replacing the~\emph{multi-head self-attentions} with large depth-wise convolutions. We mainly benchmark middle-size and large-size models, since ViTs used to be believed to surpass CNNs on large data and models. On ImageNet classification, our baseline (similar model size with \emph{Swin-B}), whose kernel size is as large as 31$\times$31, achieves \textbf{84.8\%} top-1 accuracy trained only on ImageNet-1K dataset, which is 0.3\% better than Swin-B but much more efficient in latency.

	More importantly, we find that the large kernel design is particularly powerful on \emph{downstream} tasks. For example, our networks outperform \emph{ResNeXt}-101~\cite{xie2017aggregated} or \emph{ResNet}-101~\cite{he2016deep} backbones by \textbf{4.4\%} on \emph{COCO} detection~\cite{lin2014microsoft} and \textbf{6.1\%} on \emph{ADE20K} segmentation~\cite{zhou2019semantic} under the similar complexity and parameter budget, which is also on par with or even better than the counterpart \emph{Swin Transformers} but with higher inference speed. Given more pretraining data (\eg, 73M images) and more computational budget, our best model obtains very competitive results among the state-of-the-arts with similar model sizes, \eg. \textbf{87.8\%} top-1 accuracy on ImageNet and \textbf{56.0\%} on ADE20K, which shows excellent scalability towards large-scale applications.

	We believe the high performance of RepLKNet is mainly because of the large \emph{effective receptive fields (ERFs)}~\cite{erf} built via large kernels, as compared in Fig.~\ref{fig-rf}. Moreover, RepLKNet is shown to leverage more shape information than conventional CNNs, which partially agrees with human's cognition. We hope our findings can help to understand the intrinsic mechanism of both CNNs and ViTs.

	\begin{table*}[t]
	\caption{Inference speed of a stack of 24-layer depth-wise convolutions with various kernel sizes and resolutions on a single GTX 2080Ti GPU. The input shape is (64, 384, $R$, $R$). Baselines are evaluated with Pytorch 1.9.0 + cuDNN 7.6.5, in FP32 precision.}
	\label{table-speed-kernelsize}
	\small
	\begin{center}
	\vspace{-0.2in}
    \begin{tabular}{llccccccccccccc}
    \hline
    \multirow{2}{*}{Resolution $R$} & \multirow{2}{*}{Impl} & \multicolumn{10}{c}{Latency (ms) @ Kernel size} \\
                                &                           
                                & 3     & 5     & 7     & 9     & 13   & 17     & 21    & 27    & 29    & 31        \\ \hline
    \multirow{2}{*}{$16\times 16$}         & Pytorch        
                                & 5.6   & 11.0  & 14.4  & 17.6  & 36.0 & 57.2   & 83.4  & 133.5 & 150.7 & 171.4       \\
                                & Ours                      
                                & 5.6   & 6.5   & 6.4   & 6.9   & 7.5  & 8.4    & 8.4   & 8.4   & 8.3   & 8.4       \\ \hline
    \multirow{2}{*}{$32\times 32$}         & Pytorch        
                                & 21.9  & 34.1  & 54.8  & 76.1  & 141.2 & 230.5 & 342.3 & 557.8 & 638.6 & 734.8       \\
                                & Ours                      
                                & 21.9  & 28.7  & 34.6  & 40.6  & 52.5  & 64.5  & 73.9  & 87.9  & 92.7  & 96.7       \\ \hline
    \multirow{2}{*}{$64\times 64$}         & Pytorch                   
                                & 69.6  & 141.2 & 228.6 & 319.8 & 600.0 & 977.7 & 1454.4 & 2371.1 & 2698.4 & 3090.4      \\
                                & Ours  
                                & 69.6  & 112.6 & 130.7 & 152.6 & 199.7 & 251.5 & 301.0 & 378.2 & 406.0 & 431.7       \\ \hline
    \end{tabular}
    \end{center}
    \vspace{-0.3in}
    \end{table*}

	\section{Related Work}
	
	\subsection{Models with Large Kernels}
	
	As mentioned in the introduction, apart from a few old-fashioned models like \emph{Inceptions}~\cite{szegedy2015going,szegedy2016rethinking,szegedy2017inception}, large-kernel models became not popular after \emph{VGG-Net}~\cite{simonyan2014very}. One representative work is \emph{Global Convolution Networks (GCNs)}~\cite{peng2017large}, which uses very large convolutions of 1$\times$K followed by K$\times$1 to improve semantic segmentation task. However, large kernels are reported to harm the performance on ImageNet. \emph{Local Relation Networks (LR-Net)}~\cite{hu2019local} proposes a spatial aggregation operator (LR-Layer) to replace standard convolutions, which can be viewed as a dynamic convolution. LR-Net could benefit from a kernel size of 7$\times$7, but the performance decreases with 9$\times$9. With a kernel size as large as the feature map, the top-1 accuracy significantly reduced from 75.7\% to 68.4\%. 
	
	Recently, \emph{Swin Transformers}~\cite{swin} propose to capture the spatial patterns with shifted window attention, whose window sizes range from 7 to 12, which can also be viewed as a variant of large kernel. The follow-ups \cite{dong2021cswin,liu2021swin} employ even larger window sizes. Inspired by the success of those local transformers, a recent work~\cite{han2021demystifying} replaces \emph{MHSA} layers with static or dynamic 7$\times$7 depth-wise convolutions in \cite{swin} while still maintains comparable results. Though the network proposed by \cite{han2021demystifying} shares similar design pattern with ours, the motivations are different: \cite{han2021demystifying} does not investigate the relationship between \emph{ERFs}, large kernels and performances; instead, it attributes the superior performances of vision transformers to sparse connections, shared parameters and dynamic mechanisms. Another three representative works are \emph{Global Filter Networks (GFNets)}~\cite{rao2021global}, \emph{CKConv}~\cite{romero2021ckconv} and \emph{FlexConv}~\cite{romero2021flexconv}. GFNet optimizes the spatial connection weights in the Fourier domain, which is equivalent to circular global convolutions in the spatial domain. CKConv formulates kernels as continuous functions to process sequential data, which can construct arbitrarily large kernels. FlexConv learns different kernel sizes for different layers, which can be as large as the feature maps. Although they use very large kernels, they do not intend to answer the key questions we desire: why do traditional CNNs underperform ViTs, and how to apply large kernels in \emph{common CNNs}.	Besides, both \cite{han2021demystifying} and \cite{rao2021global} do not evaluate their models on strong baselines, \eg, models larger than \emph{Swin-L}. Hence it is still unclear whether large-kernel CNNs can scale up well as transformers.

	\noindent \textbf{Concurrent works}.  
	
	\emph{ConvMixer} \cite{trockman2022patches} uses up to 9$\times$9 convolutions to replace the ``mixer'' component of \emph{ViTs}~\cite{vit} or \emph{MLPs}~\cite{tolstikhin2021mlp,touvron2021resmlp}. \emph{MetaFormer}~\cite{yu2021metaformer} suggests \emph{pooling} layer is an alternate to self-attention. \emph{ConvNeXt}~\cite{liu2022convnet} employs 7$\times$7 depth-wise convolutions to design strong architectures, pushing the limit of CNN performances. Although those works show excellent performances, they do not show benefits from much larger convolutions (\eg, 31$\times$31).

	\subsection{Model Scaling Techniques}
	
	Given a small model, it is a common practice to scale it up for better performance, thus scaling strategy plays a vital role in the resultant accuracy-efficiency trade-offs. For CNNs, existing scaling approaches usually focus on model depth, width, input resolution~\cite{efficientnet,regnet,dollar2021fast}, bottleneck ratio and group width~\cite{regnet,dollar2021fast}. Kernel size, however, is often neglected. In Sec.~\ref{sect-3}, we will show that the kernel size is also an important scaling dimension in CNNs, especially for \emph{downstream} tasks.

	\subsection{Structural Re-parameterization}
	
	Structural Re-parameterization~\cite{ding2021repvgg,ding2021diverse,ding2021resrep,ding2021repmlpnet,ding2019acnet} is a methodology of equivalently converting model structures via transforming the parameters. For example, RepVGG targeted at a deep inference-time VGG-like (\eg, branch-free) model, and constructed extra ResNet-style shortcuts parallel to the 3$\times$3 layers during training. In contrast to a real VGG-like model that is difficult to train~\cite{he2016deep}, such shortcuts helped the model reach a satisfactory performance. After training, the shortcuts are absorbed into the parallel 3$\times$3 kernels via a series of linear transformations, so that the resultant model becomes a VGG-like model. In this paper, we use this methodology to add a relatively small (\eg, 3$\times$3 or 5$\times$5) kernel into a very large kernel. In this way, we make the very large kernel capable of capturing small-scale patterns, hence improve the performance of the model.

	\section{Guidelines of Applying Large Convolutions}\label{sect-3}
	
	Trivially applying large convolutions to CNNs usually leads to inferior performance and speed. In this section, we summarize 5 guidelines for effectively using large kernels.
	
	\vspace{-0.15in}\paragraph{Guideline 1: large depth-wise convolutions can be efficient in practice.} 
	It is believed that large-kernel convolutions are computationally expensive because the kernel size quadratically increases the number of parameters and FLOPs. The drawback can be greatly overcome by applying depth-wise (DW) convolutions \cite{mbv1,chollet2017xception}. For example, in our proposed \emph{RepLKNet} (see Table~\ref{table-replknet-224} for details), increasing the kernel sizes in different stages from $[3,3,3,3]$ to $[31,29,27,13]$ only increases the FLOPs and number of parameters by 18.6\% and 10.4\% respectively, which is acceptable. The remaining 1$\times$1 convolutions actually dominate most of the complexity. 
	
	One may concern that DW convolutions could be very inefficient on modern parallel computing devices like GPUs. It is true for conventional DW 3$\times$3 kernels~\cite{mbv1,mbv2,zhang2018shufflenet}, because DW operations introduce low ratio of computation \vs memory access cost~\cite{ma2018shufflenet}, which is not friendly to modern computing architecture. However, we find when kernel size becomes large, the computational density increases: for example, in a DW 11$\times$11 kernel, each time we load a value from the feature map, it can attend at most 121 multiplications, while in a 3$\times$3 kernel the number is only 9. Therefore, according to the roofline model, the actual latency should not increase as much as the increasing of FLOPs when kernel size becomes larger. 
	
	\noindent \textbf{Remark 1.} Unfortunately, we find off-the-shelf deep learning tools (such as Pytorch) support large DW convolutions poorly, as shown in Table~\ref{table-speed-kernelsize}. Hence we try several approaches to optimize the CUDA kernels. FFT-based approach~\cite{mathieu2013fast} appears reasonable to implement large convolutions. However, in practice we find block-wise (inverse) \emph{implicit gemm} algorithm is a better choice. The implementation has been integrated into the open-sourced framework MegEngine~\cite{MegEngine} and we omit the details here. We have also released an efficient implementation~\cite{RepLKNet-pytorch} for PyTorch. Table~\ref{table-speed-kernelsize} shows that our implementation is far more efficient, compared with the Pytorch baseline. With our optimization, the latency contribution of DW convolutions in RepLKNet reduces from 49.5\% to 12.3\%, which is roughly in proportion to the FLOPs occupation. 
	
	\vspace{-0.15in}
	\paragraph{Guideline 2: identity shortcut is vital especially for networks with very large kernels.} 
To demonstrate this, we use \emph{MobileNet V2}~\cite{mbv2} to benchmark, since it heavily uses DW layers and has two published variants (with or without shortcuts). For the large-kernel counterparts, we simply replace all the DW 3$\times$3 layers with 13$\times$13. All the models are trained on ImageNet with the identical training configurations for 100 epochs (see Appendix A for details). Table~\ref{table-mob2-shortcut} shows large kernels improve the accuracy of MobileNet V2 with shortcuts by 0.77\%. However, without shortcuts, large kernels reduce the accuracy to only 53.98\%.

\noindent \textbf{Remark 2.} The guideline also works for \emph{ViTs}. A recent work~\cite{dong2021attention} finds that without identity shortcut, attention loses rank doubly exponentially with depth, leading to over-smoothing issue. Although large-kernel CNNs may degenerate in a different mechanism from ViT's, we also observed without shortcut, it is difficult for the network to capture local details. From a similar perspective as \cite{veit2016residual}, shortcuts make the model an implicit ensemble composed of numerous models with different receptive fields (RFs), so it can benefit from a much larger maximum RF while not losing the ability to capture small-scale patterns.

	\begin{table}
		\caption{Results of different kernel sizes in normal/shortcut-free MobileNet V2.}
		\label{table-mob2-shortcut}
		\vspace{-0.2in}
		\begin{center}
			\small
			\begin{tabular}{lcccc}
				\hline
				Shortcut 		& Kernel size		& ImageNet top-1 accuracy (\%)	\\
				\hline
				\checkmark		&	3$\times$3		&	71.76		\\
				\checkmark		&	13$\times$13	&	\textbf{72.53}		\\
		        \hline
				&	3$\times3$		&	\textbf{68.67}		\\
				&	13$\times$13	&	53.98		\\		
				\hline
			\end{tabular}
		\end{center}
		\vspace{-0.25in}
	\end{table}
	
	\begin{figure*}
		\begin{center}
			\includegraphics[width=\linewidth]{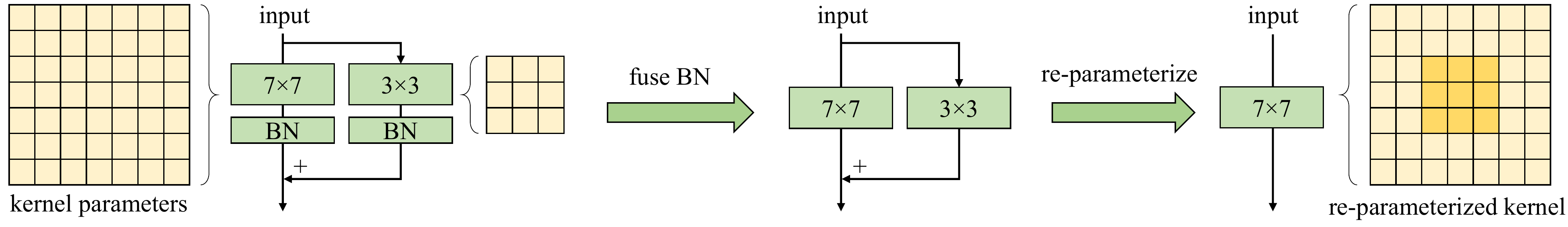}
			\vspace{-0.25in}
			\caption{An example of re-parameterizing a small kernel (\eg, 3$\times$3) into a large one (\eg, 7$\times$7). See \cite{ding2019acnet,ding2021repvgg} for details. }
			\label{fig-reparam}
			\vspace{-0.25in}
		\end{center}
	\end{figure*}
	
	\vspace{-0.15in} \paragraph{Guideline 3: re-parameterizing \cite{ding2021repvgg} with small kernels helps to make up the optimization issue.} We replace the 3$\times$3 layers of \emph{MobileNet V2} by 9$\times$9 and 13$\times$13 respectively, and optionally adopt \emph{Structural Re-parameterization}~\cite{ding2021repvgg,ding2021repmlpnet,ding2019acnet} methodology. Specifically, we construct a 3$\times$3 layer parallel to the large one, then add up their outputs after \emph{Batch normalization (BN)}~\cite{ioffe2015batch} layers (Fig.~\ref{fig-reparam}). After training, we merge the small kernel as well as BN parameters into the large kernel, so the resultant model is equivalent to the model for training but no longer has small kernels. Table~\ref{table-mob2-reparam} shows directly increasing the kernel size from 9 to 13 reduces the accuracy, while re-parameterization addresses the issue. 
	
	We then transfer the ImageNet-trained models to semantic segmentation with DeepLabv3+~\cite{chen2018encoder} on Cityscapes~\cite{cityscapes}. We only replace the backbone and keep all the default training settings provided by MMSegmentation~\cite{mmseg2020}. The observation is similar to that on ImageNet: 3$\times$3 re-param improves the mIoU of the 9$\times$9 model by 0.19 and the 13$\times$13 model by 0.93. With such simple re-parameterization, increasing kernel size from 9 to 13 no longer degrades the performance on both ImageNet and Cityscapes. 
	
	\noindent\textbf{Remark 3.} 
	It is known that \emph{ViTs} have optimization problem especially on small datasets~\cite{vit,liu2020understanding}. A common workaround is to introduce \emph{convolutional prior}, \eg, add a DW 3$\times$3 convolution to each self-attention block~\cite{wu2021cvt,chu2021conditional}, which is analogous to ours. Those strategies introduce additional \emph{translational equivariance} and \emph{locality} prior to the network, making it easier to optimize on small dataset without loss of generality. Similar to what ViT behaves~\cite{vit}, we also find when the pretraining dataset increases to 73 million images (refer to \emph{RepLKNet-XL} in the next section), re-parameterization can be omitted without degradation. 
	
	\begin{table}
		\caption{Results of 3$\times$3 re-parameterization on MobileNet V2 with various kernel sizes.}
		\label{table-mob2-reparam}
		\vspace{-0.2in}
		\begin{center}
			\small
			\begin{tabular}{lcccc}
				\hline
				Kernel 					&	3$\times$3 re-param		& \makecell{ImageNet \\ top-1 acc (\%)}	&	\makecell{Cityscapes \\ val mIoU (\%)}	\\
				\hline
				3$\times$3				&  N/A						&	71.76				&	72.31	\\
				\hline
				9$\times$9			&						&	72.67				&	76.11	\\
				9$\times$9			&	\checkmark			&	\textbf{73.09}				&	\textbf{76.30}	\\
				\hline
				13$\times$13		&						&	72.53				&	75.67	\\
				13$\times$13		&	\checkmark			&	\textbf{73.24}				&	\textbf{76.60}	\\			
				\hline
			\end{tabular}
		\end{center}
		\vspace{-0.25in}
	\end{table}

	\vspace{-0.15in}\paragraph{Guideline 4: large convolutions boost downstream tasks much more
than ImageNet classification.} 
Table~\ref{table-mob2-reparam} (after \emph{re-param}) shows increasing the kernel size of MobileNet V2 from 3$\times$3 to 9$\times$9 improves the \emph{ImageNet} accuracy by 1.33\% but the \emph{Cityscapes} mIoU by 3.99\%. Table~\ref{table-replknet-224} shows a similar trend: as the kernel sizes increase from $[3,3,3,3]$ to $[31,29,27,13]$, the ImageNet accuracy improves by only 0.96\%, while the mIoU on \emph{ADE20K}~\cite{zhou2019semantic} improves by 3.12\%. Such phenomenon indicates that models of similar ImageNet scores could have very different capability in downstream tasks (just as the bottom 3 models in Table~\ref{table-replknet-224}).  

\noindent \textbf{Remark 4.} 
What causes the phenomenon? First, large kernel design significantly increases the \emph{Effective Receptive Fields (ERFs)} \cite{erf}. Numerous works have demonstrated ``contextual'' information, which implies large ERFs, is crucial in many downstream tasks like object detection and semantic segmentation \cite{peng2017large,long2015fully,yu2017dilated,wang2020deep,yu2015multi}. We will discuss the topic in Sec.~\ref{sec:largeerf}. 
Second, We deem another reason might be that large kernel design contributes more shape biases to the network. Briefly speaking, ImageNet pictures can be correctly classified according to either texture or shape, as proposed in \cite{geirhos2018imagenet,brendel2019approximating}. However, humans recognize objects mainly based on shape cue rather than texture, therefore a model with stronger shape bias may transfer better to downstream tasks. A recent study \cite{tuli2021convolutional} points out ViTs are strong in shape bias, which partially explains why ViTs are super powerful in transfer tasks. In contrast, conventional CNNs trained on ImageNet tend to bias towards texture \cite{geirhos2018imagenet,brendel2019approximating}. Fortunately, we find simply enlarging the kernel size in CNNs can effectively improve the shape bias. Please refer to Appendix C for details.
	
	\begin{figure}
		\begin{center}
			\includegraphics[width=\linewidth]{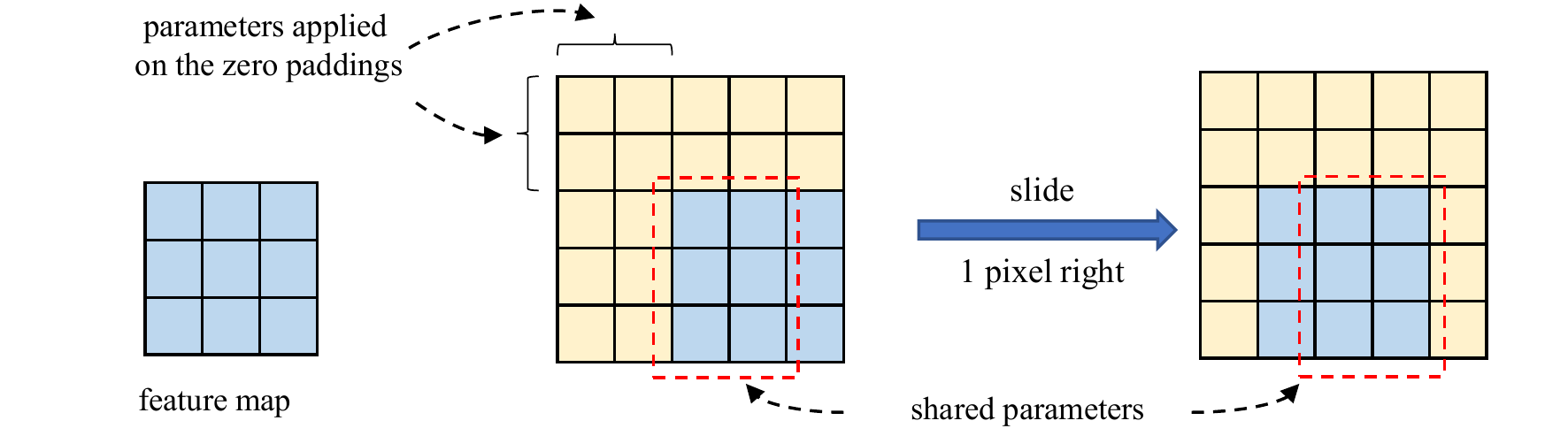}
			\vspace{-0.25in}
			\caption{Illustration to convolution with small feature map and large kernel. Two outputs at adjacent locations only share a part of kernel weights. Translational equivariance does not strictly hold.}
			\label{fig-large-kernel-small-feature}
			\vspace{-0.15in}
		\end{center}
	\end{figure}
	
	\vspace{-0.15in}\paragraph{Guideline 5: large kernel (\eg, 13\bm{$\times$}13) is useful even on small feature maps (\eg, 7\bm{$\times$}7).} 
	To validate it, We enlarge the DW convolutions in the \emph{last stage} of \emph{MobileNet V2} to 7$\times$7 or 13$\times$13, hence the kernel size is on par with or even larger than feature map size (7$\times$7 by default). We apply re-parameterization to the large kernels as suggested by Guideline 3. Table~\ref{table-mob2-smallfeature} shows although convolutions in the last stage already involve very large receptive field, further increasing the kernel sizes still leads to performance improvements, especially on downstream tasks such as \emph{Cityscapes}. 
	
	\noindent \textbf{Remark 5.} 
	When kernel size becomes large, notice that translational equivariance of CNNs does not strictly hold. As illustrated in Fig.~\ref{fig-large-kernel-small-feature}, two outputs at adjacent spatial locations share only a fraction of the kernel weights, \ie, are transformed by different mappings. The property also agrees with the ``philosophy'' of \emph{ViTs} -- relaxing the symmetric prior to obtain more capacity. Interestingly, we find 2D \emph{Relative Position Embedding (RPE)}~\cite{shaw2018self,bello2019attention}, which is widely used in the transformer community, can also be viewed as a large depth-wise kernel of size $(2H-1)\times(2W-1)$, where $H$ and $W$ are feature map height and width respectively. Large kernels not only help to learn the relative positions between concepts, but also encode the \emph{absolute position} information due to \emph{padding effect}~\cite{kayhan2020translation}. 
	
	\begin{table}
		\caption{Results of various kernel sizes in the \emph{last stage} of MobileNet V2. Kernel sizes in previous stages remain to be $3\times 3$.}
		\label{table-mob2-smallfeature}
		\vspace{-0.2in}
		\begin{center}
			\small
			\begin{tabular}{lcccc}
				\hline
				Kernel size		& ImageNet acc (\%)	&	Cityscapes mIoU (\%)	\\
				\hline
				3$\times$3		&	71.76				&	72.31	\\
				7$\times$7		&	\textbf{72.00}				&	74.30	\\
				13$\times$13	&	71.97				&	\textbf{74.62}	\\		
				\hline
			\end{tabular}
			
		\end{center}
		\vspace{-0.2in}
	\end{table}

	\section{RepLKNet: a Large-Kernel Architecture}
	\label{sec:arch}
	Following the above guidelines, in this section we propose RepLKNet, a \emph{pure} CNN architecture with large kernel design. To our knowledge, up to now CNNs still dominate small models~\cite{zhang2021collaboration,zhang2021basisnet}, while \emph{vision transformers} are believed to be better than CNNs under more complexity budget. Therefore, in the paper we mainly focus on relatively large models (whose complexity is on par with or larger than \emph{ResNet-152}~\cite{he2016deep} or \emph{Swin-B}~\cite{swin}), in order to verify whether large kernel design could eliminate the performance gap between CNNs and ViTs.

	\subsection{Architecture Specification}
	
\begin{figure}
		\begin{center}
			\includegraphics[width=\linewidth]{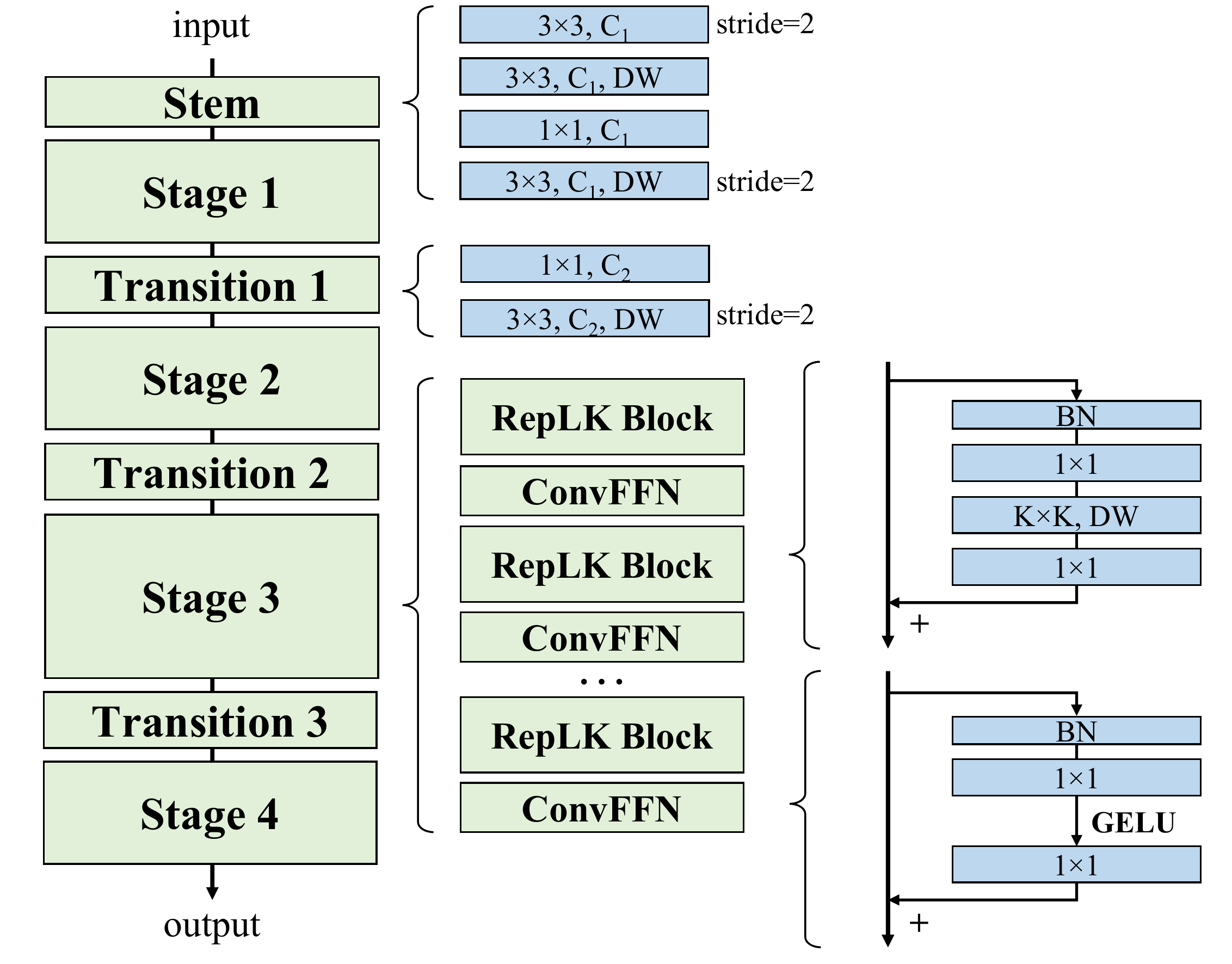}
			\vspace{-0.3in}
			\caption{RepLKNet comprises Stem, Stages and Transitions. Except for depth-wise (DW) large kernel, the other components include DW 3$\times$3, dense 1$\times$1 conv, and batch normalization~\cite{ioffe2015batch} (BN). Note that every conv layer has a following BN, which are not depicted. Such conv-BN sequences use ReLU as the activation function, except those before the shortcut-addition (as a common practice \cite{he2016deep,mbv2}) and those preceding GELU~\cite{hendrycks2016gaussian}.}
			\label{fig-lknet-arch}
			\vspace{-0.2in}
		\end{center}
	\end{figure}

	We sketch the architecture of RepLKNet in Fig.~\ref{fig-lknet-arch}:
	
	\textbf{Stem} refers to the beginning layers. Since we target at high performance on downstream dense-prediction tasks, we desire to capture more details by several conv layers at the beginning. After the first 3$\times$3 with 2$\times$ downsampling, we arrange a DW 3$\times$3 layer to capture low-level patterns, a 1$\times$1 conv, and another DW 3$\times$3 layer for downsampling. 
	
	\textbf{Stages} 1-4 each contains several RepLK Blocks, which use shortcuts (Guideline 2) and DW large kernels (Guideline 1). We use 1$\times$1 conv before and after DW conv as a common practice. Note that each DW large conv uses a 5$\times$5 kernel for re-parameterization (Guideline 3), which is not shown in Fig.~\ref{fig-lknet-arch}. Except for the large conv layers which provide sufficient receptive field and the ability to aggregate spatial information, the model's representational capacity is also closely related to the depth. To provide more nonlinearities and information communications across channels, we desire to use 1$\times$1 layers to increase the depth. Inspired by the Feed-Forward Network (FFN) which has been widely used in transformers~\cite{vit,swin} and MLPs~\cite{tolstikhin2021mlp,touvron2021resmlp,ding2021repmlpnet}, we use a similar CNN-style block composed of shortcut, BN, two 1$\times$1 layers and GELU~\cite{hendrycks2016gaussian}, so it is referred to as ConvFFN Block. Compared to the classic FFN which uses Layer Normalization~\cite{ba2016layer} before the fully-connected layers, BN has an advantage that it can be fused into conv for efficient inference. As a common practice, the number of internal channels of the ConvFFN Block is 4$\times$ as the input. Simply following ViT and Swin, which interleave attention and FFN blocks, we place a ConvFFN after each RepLK Block.
	
	\textbf{Transition Blocks} are placed between stages, which first increase the channel dimension via 1$\times$1 conv and then conduct 2$\times$ downsampling with DW 3$\times$3 conv.
	
	In summary, each stage has three architectural hyper-parameters: the number of RepLK Blocks $B$, the channel dimension $C$, and the kernel size $K$. So that a RepLKNet architecture is defined by $[B_1,B_2,B_3,B_4]$,$[C_1,C_2,C_3,C_4]$,$[K_1,K_2,K_3,K_4]$.
	
	\subsection{Making Large Kernels Even Larger}\label{sect-replknet-120}
	
	\setlength{\tabcolsep}{3pt}
	\begin{table}
		\caption{RepLKNet with different kernel sizes. The models are pretrained on ImageNet-1K in 120 epochs with 224$\times$224 input and finetuned on ADE20K with UperNet in 80K iterations. On ADE20K, we test the \emph{single-scale} mIoU, and compute the FLOPs with input of 2048$\times$512, following Swin.}
		\label{table-replknet-224}
		\vspace{-0.25in}
		\begin{center}		
			\small
			\begin{tabular}{l|ccc|ccc}
				\hline
				& \multicolumn{3}{c|}{ImageNet} & \multicolumn{3}{c}{ADE20K} \\
				Kernel size		&	Top-1		& \makecell{Params}		&\makecell{FLOPs}	&	mIoU	& Params	& FLOPs	\\
				\hline
				3-3-3-3         &   82.11           &   71.8M       &   12.9G   &   46.05   &   104.1M  &   1119G\\
				7-7-7-7         &   82.73           &   72.2M       &   13.1G   &   48.05   &   104.6M  &   1123G\\
				13-13-13-13		&	83.02			&	73.7M		&	13.4G	&   48.35   &   106.0M  &   1130G\\ 
				25-25-25-13		&	83.00			&	78.2M       &   14.8G	&   48.68   &   110.6M  &   1159G\\
				31-29-27-13     &   83.07           &   79.3M       &   15.3G   &   49.17   &   111.7M  &   1170G\\
				\hline
			\end{tabular}
		\end{center}
		\vspace{-0.3in}
	\end{table}
	\setlength{\tabcolsep}{1.4pt}
	
	We continue to evaluate large kernels on RepLKNet via fixing $\vect{B}$=$[2,2,18,2]$, $\vect{C}$=$[128,256,512,1024]$, varying $\vect{K}$ and observing the performance of both classification and semantic segmentation. Without careful tuning of the hyper-parameters, we casually set the kernel sizes as $[13,13,13,13]$, $[25,25,25,13]$, $[31,29,27,13]$, respectively, and refer to the models as RepLKNet-13/25/31. We also construct two small-kernel baselines where the kernel sizes are all 3 or 7 (RepLKNet-3/7). 
	
	On ImageNet, we train for 120 epochs with AdamW~\cite{loshchilov2017decoupled} optimizer, RandAugment~\cite{cubuk2020randaugment}, mixup~\cite{zhang2017mixup}, CutMix~\cite{yun2019cutmix}, Rand Erasing~\cite{zhong2020random} and Stochastic Depth~\cite{stochastic}, following the recent works~\cite{swin,deit,bao2021beit,liu2022convnet}. The detailed training configurations are presented in Appendix A.
	
	For semantic segmentation, we use ADE20K~\cite{zhou2019semantic}, which is a widely-used large-scale semantic segmentation dataset containing 20K images of 150 categories for training and 2K for validation. We use the ImageNet-trained models as backbones and adopt UperNet~\cite{xiao2018unified} implemented by MMSegmentation~\cite{mmseg2020} with the 80K-iteration training setting and test the \emph{single-scale} mIoU.
	
	Table~\ref{table-replknet-224} shows our results with different kernel sizes. On ImageNet, though increasing the kernel sizes from 3 to 13 improves the accuracy, making them even larger brings no further improvements. However, on ADE20K, scaling up the kernels from $[13,13,13,13]$ to $[31,29,27,13]$ brings 0.82 higher mIoU with only 5.3\% more parameters and 3.5\% higher FLOPs, which highlights the significance of large kernels for downstream tasks.
	
	In the following subsections, we use RepLKNet-31 with stronger training configurations to compare with the state-of-the-arts on ImageNet classification, Cityscapes/ADE20K semantic segmentation and COCO~\cite{lin2014microsoft} object detection. We refer to the aforementioned model as RepLKNet-31B (B for Base) and a wider model with $\vect{C}=[192,384,768,1536]$ as RepLKNet-31L (Large). We construct another RepLKNet-XL with $\vect{C}=[256,512,1024,2048]$ and 1.5$\times$ inverted bottleneck design in the RepLK Blocks (\ie, the channels of the DW large conv layers are 1.5$\times$ as the inputs).
	
	\setlength{\tabcolsep}{2pt}
	\begin{table}
		\caption{ImageNet results. The throughput is tested with FP32 and a batch size of 64 on 2080Ti. $\ddagger$ indicates ImageNet-22K pretraining. $\diamond$ indicates pretrained with extra data.}
		\label{table-to-swin}
		\vspace{-0.25in}
		\begin{center}		
			\small
			\begin{tabular}{llcccccc}
				\hline
				Model			& \makecell{Input \\ resolution}& \makecell{Top-1 \\ acc}	&	\makecell{Params \\(M)}	& \makecell{FLOPs \\(G)} & \makecell{Throughput\\examples/s}\\
				\hline
				\textbf{RepLKNet-31B}     &	224$\times$224 & 	83.5            &   79   & 15.3 & 295.5\\
				Swin-B     	&	224$\times$224&   	83.5    		&	88		& 15.4	    &  226.2 \\
				\hline
				\textbf{RepLKNet-31B}     &	384$\times$384& 	84.8            &   79   	& 45.1 & 97.0   \\
				Swin-B     	&	384$\times$384&   	84.5    		&	88		& 47.0	    & 67.9\\
				\hline
				\textbf{RepLKNet-31B}~$^\ddagger$     &	224$\times$224 & 	85.2            &   -  & - & - \\
				Swin-B~$^\ddagger$     	&	224$\times$224&   	85.2    		&	-		& -	    &  - \\
				\hline
				\textbf{RepLKNet-31B}~$^\ddagger$     &	384$\times$384& 	86.0            &   -   	& - & -   \\
				Swin-B~$^\ddagger$     	&	384$\times$384&   	86.4    		&	-		& -	    & -\\
				\hline
				\textbf{RepLKNet-31L}~$^\ddagger$   &	384$\times$384& 	86.6  &   172 &   96.0 &   50.2               \\
				Swin-L~$^\ddagger$ 	       &	384$\times$384&    87.3    &    197 &103.9 &   36.2 	                \\
				
				\hline
				\textbf{RepLKNet-XL}~$^\diamond$ & 320$\times$320 & 87.8 & 335 & 128.7 & 39.1 \\
				\hline
			\end{tabular}
		\end{center}
		\vspace{-0.25in}
	\end{table}

	\setlength{\tabcolsep}{2pt}
	\begin{table}
		\caption{Cityscapes results. The FLOPs is computed with 1024$\times$2048 inputs. The mIoU is tested with single-scale (ss) and multi-scale (ms). The results with Swin are implemented by~\cite{gong2021improve}. $\ddagger$ indicates ImageNet-22K pretraining.}
		\label{table-compare-cityscapes}
		\vspace{-0.25in}
		\begin{center}		
			\small
			\begin{tabular}{lllccccc}
				\hline
				Backbone			&	Method		&		\makecell{mIoU \\ (ss)}	&	\makecell{mIoU \\ (ms)}	& 	\makecell{Param \\ (M)}	& \makecell{FLOPs \\ (G)}\\
				\hline
				\textbf{RepLKNet-31B}     	&	UperNet \cite{xiao2018unified}		&		\textbf{83.1}		&	\textbf{83.5}			&	110	&	2315\\			
				ResNeSt-200~\cite{resnest}         &   DeepLabv3~\cite{chen2017rethinking}   &       -           &   82.7            &   -       &   -   \\
				Axial-Res-XL	&Axial-DL \cite{wang2020axial}		&	80.6		&	81.1		&	173	&	2446	\\
				Swin-B					&	UperNet		&		80.4		&	81.5		&	121		&	2613	\\	Swin-B &	UperNet + \cite{gong2021improve}		&		80.8		&	81.8		&	121		&	-	\\
				\hline
				ViT-L~$^\ddagger$	&	SETR-PUP \cite{zheng2021rethinking}	&		79.3		&	82.1		&	318	&	-	\\
				ViT-L~$^\ddagger$	&	SETR-MLA &		77.2		&	-			&	310	&	-		\\
				Swin-L~$^\ddagger$	&	UperNet		&		82.3		&	83.1		&	234		&	3771	\\
				Swin-L~$^\ddagger$     &	UperNet + \cite{gong2021improve}		&		82.7		&	83.6		&	234		&	-	\\
				\hline
			\end{tabular}
		\end{center}
		\vspace{-0.3in}
	\end{table}

	\setlength{\tabcolsep}{2pt}
	\begin{table}
		\caption{ADE20K results. The mIoU is tested with single-scale (ss) and multi-scale (ms). The results with 1K-pretrained Swin are cited from the official GitHub repository. $\ddagger$ indicates ImageNet-22K pretraining and 640$\times$640 finetuning on ADE20K. $\diamond$ indicates pretrained with extra data. The FLOPs is computed with 2048$\times$512 for the ImageNet-1K pretrained models and 2560$\times$640 for the ImageNet-22K and larger, following Swin.}
		\label{table-compare-ade20K}
		\vspace{-0.25in}
		\begin{center}		
			\small
			\begin{tabular}{lllcccc}
				\hline
					Backbone			&	Method		&		\makecell{mIoU \\ (ss)}	&	\makecell{mIoU \\ (ms)}	& 	\makecell{Param \\ (M)}	& \makecell{FLOPs \\ (G)}\\
				\hline
				\textbf{RepLKNet-31B}   	&	UperNet		&		\textbf{49.9}		&	\textbf{50.6}			&	112	&	1170\\   
				ResNet-101				&	UperNet \cite{xiao2018unified}			&	43.8	&	44.9		&	86		&	1029	\\
				ResNeSt-200~\cite{resnest}     &	DeepLabv3~\cite{chen2017rethinking}	&	-	&	48.4		&	113		&	1752	\\
				Swin-B	&	UperNet			&		48.1		&	49.7		&	121	&	1188	\\
				Swin-B           &	UperNet + \cite{gong2021improve}		&	48.4	&	50.1	&	121	&	-	\\
				ViT-Hybrid			&DPT-Hybrid~\cite{ranftl2021vision}				&		-			&	49.0		&	90		&	-	\\
				ViT-L				&	DPT-Large			&		-			&	47.6		&	307	&	-	\\
				ViT-B					&	SETR-PUP \cite{zheng2021rethinking}	&		46.3		&	47.3		&	97		&	-	\\
				ViT-B						&	SETR-MLA \cite{zheng2021rethinking}	&		46.2		&	47.7		&	92		&	-		\\
				\hline
				
				\textbf{RepLKNet-31B}~$^\ddagger$   	&	UperNet		&		\textbf{51.5}		&	\textbf{52.3}			&	112	&	1829\\
				Swin-B~$^\ddagger$	&	UperNet			&		50.0		&	51.6		&	121	&	1841	\\

				\textbf{RepLKNet-31L}~$^\ddagger$   	&	UperNet		&		\textbf{52.4}		&	52.7			&	207	&	2404\\	
				Swin-L~$^\ddagger$	&	UperNet			&		52.1		&	\textbf{53.5}		&	234	    &	2468	\\
				ViT-L~$^\ddagger$		&	SETR-PUP	&		48.6		&	50.1		&	318		&	-	\\
				ViT-L~$^\ddagger$		&	SETR-MLA	&		48.6		&	50.3		&	310		&	-		\\
				\hline
				\textbf{RepLKNet-XL}~$^\diamond$ & UperNet & \textbf{55.2} & \textbf{56.0} & 374 & 3431 \\
				\hline
			\end{tabular}
		\end{center}
		\vspace{-0.3in}
	\end{table}

	\subsection{ImageNet Classification}\label{sect-imgnet}
	
	Since the overall architecture of RepLKNet is akin to Swin, we desire to make a comparison at first. For RepLKNet-31B on ImageNet-1K, we extend the aforementioned training schedule to 300 epochs for a fair comparison. Then we finetune for 30 epochs with input resolution of 384$\times$384, so that the total training cost is much lower than the Swin-B model, which was trained with 384$\times$384 from scratch. Then we pretrain RepLKNet-B/L models on ImageNet-22K and finetune on ImageNet-1K. RepLKNet-XL is pretrained on our private semi-supervised dataset named \emph{MegData73M}, which is introduced in the Appendix. We also present the throughput tested with a batch size of 64 on the same 2080Ti GPU. The training configurations are presented in the Appendix.
	
	Table~\ref{table-to-swin} shows that though very large kernels are not intended for ImageNet classification, our RepLKNet models show a a favorable trade-off between accuracy and efficiency. Notably, with only ImageNet-1K training, RepLKNet-31B reaches 84.8\% accuracy, which is 0.3\% higher than Swin-B, and runs 43\% faster. And even though RepLKNet-XL has higher FLOPs than Swin-L, it runs faster, which highlights the efficiency of very large kernels.

	\subsection{Semantic Segmentation}
	
	We then use the pretrained models as the backbones on Cityscapes (Table~\ref{table-compare-cityscapes}) and ADE20K (Table~\ref{table-compare-ade20K}). Specifically, we use the UperNet~\cite{xiao2018unified} implemented by MMSegmentation~\cite{mmseg2020} with the 80K-iteration training schedule for Cityscapes and 160K for ADE20K. Since we desire to evaluate the backbone only, we do not use any advanced techniques, tricks, nor custom algorithms. 
	
	On Cityscapes, ImageNet-1K-pretrained RepLKNet-31B outperforms Swin-B by a significant margin (single-scale mIoU of 2.7), and even {outperforms the ImageNet-22K-pretrained Swin-L}. Even equipped with DiversePatch~\cite{gong2021improve}, a technique customized for vision transformers, the single-scale mIoU of the 22K-pretrained Swin-L is still lower than our 1K-pretrained RepLKNet-31B, though the former has 2$\times$ parameters.
	
	On ADE20K, RepLKNet-31B outperforms Swin-B with both 1K and 22K pretraining, and the margins of single-scale mIoU are particularly significant. Pretrained with our semi-supervised dataset \emph{MegData73M}, RepLKNet-XL achieves an mIoU of 56.0, which shows feasible scalability towards large-scale vision applications.

	\subsection{Object Detection}
	
	For object detection, we use RepLKNets as the backbone of FCOS~\cite{fcos} and Cascade Mask R-CNN~\cite{he2017mask,cai2019cascade}, which are representatives of one-stage and two-stage detection methods, and the default configurations in MMDetection~\cite{mmdetection}. The FCOS model is trained with the 2x (24-epoch) training schedule for a fair comparison with the X101 (short for ResNeXt-101~\cite{xie2017aggregated}) baseline from the same code base~\cite{mmseg2020}, and the other results with Cascade Mask R-CNN all use 3x (36-epoch). Again, we simply replace the backbone and do not use any advanced techniques. Table~\ref{table-coco} shows RepLKNets outperform ResNeXt-101-64x4d by up to 4.4 mAP while have fewer parameters and lower FLOPs. Note that the results may be further improved with the advanced techniques like HTC~\cite{chen2019hybrid}, HTC++~\cite{swin}, Soft-NMS~\cite{bodla2017soft} or a 6x (72-epoch) schedule. Compared to Swin, RepLKNets achieve higher or comparable mAP with fewer parameters and lower FLOPs. Notably, RepLKNet-XL achieves an mAP of 55.5, which demonstrates the scalability again.
	
	\setlength{\tabcolsep}{2pt}
	\begin{table}
		\caption{Object detection on COCO. The FLOPs is computed with 1280$\times$800 inputs. The results of ResNeXt-101-64x4d + Cas Mask are reported by \cite{swin}. The results of 22K-pretrained Swin (without HTC++~\cite{swin}) are reported by \cite{liu2022convnet}. $\ddagger$ indicates ImageNet-22K pretraining. $\diamond$ indicates pretrained with extra data.}
		\label{table-coco}
		\vspace{-0.25in}
		\begin{center}		
			\small
			\begin{tabular}{llcccccc}
				\hline
				Backbone				&	Method	&	$\text{AP}^{\text{box}}$ &	$\text{AP}^{\text{mask}}$   &\makecell{Param \\ (M)}	& \makecell{FLOPs \\ (G)}		\\
				\hline
				\textbf{RepLKNet-31B}    &   FCOS    &   \textbf{47.0}   &   -   &   87  &   437\\
				X101-64x4d                  &   FCOS    &    42.6   &   -   &   90  &   439 \\
				\hline
				\textbf{RepLKNet-31B}	&	Cas Mask 	&	\textbf{52.2}    &   \textbf{45.2}   &   137    &   965    \\
				X101-64x4d		&	Cas Mask 	&	48.3    &   41.7   &  140		&	972\\
				ResNeSt-200		&	Cas R-CNN \cite{cai2019cascade}	&	49.0    &   -	&	- & -	\\				
				Swin-B      	&	Cas Mask 	&	51.9    &   45.0    &   145    &   982    \\
				\hline
				\textbf{RepLKNet-31B}~$^\ddagger$	&	Cas Mask 	&	\textbf{53.0}    &   \textbf{46.0}   &   137    &   965       \\
				Swin-B~$^\ddagger$	                &	Cas Mask 	&	\textbf{53.0}    &   45.8            &   145    &   982\\
				\hline
				\textbf{RepLKNet-31L}~$^\ddagger$ & Cas Mask  & \textbf{53.9}    & 46.5  &   229 &   1321\\
				Swin-L~$^\ddagger$     & Cas Mask  & \textbf{53.9}  &   \textbf{46.7}    &   254   &   1382  \\
				\hline
				\textbf{RepLKNet-XL}~$^\diamond$ & Cas Mask & \textbf{55.5} & \textbf{48.0} &   392 &   1958 \\ 
				\hline
			\end{tabular}
		\end{center}
		\vspace{-0.35in}
	\end{table}

	\section{Discussions}
	\label{sec:largeerf}
	
	\subsection{Large-Kernel CNNs have Larger ERF than Deep Small-Kernel Models}
	
	We have demonstrated large kernel design can significantly boost \emph{CNNs} (especially on \emph{downstream} tasks). However, it is worth noting that large kernel can be expressed by a series of small convolutions \cite{simonyan2014very}, \eg, a 7$\times$7 convolution can be decomposed into a stack of three 3$\times$3 kernels without information loss (more channels are required after the decomposition to maintain the degree of freedom). Given that fact, a question naturally comes up: why do conventional CNNs, which may contain tens or hundreds of small convolutions (\eg, \emph{ResNets}~\cite{he2016deep}), still behave inferior to large-kernel networks?
	
	We argue that in terms of obtaining large receptive field, a single large kernel is much more \emph{effective} than many small kernels. \textbf{First}, according to the theory of \emph{Effective Receptive Field (ERF)}~\cite{erf}, ERF is proportion to $\mathcal{O}(K\sqrt{L})$, where $K$ is the kernel size and $L$ is the depth, \ie, number of layers. In other words, ERF grows linearly with the kernel size while sub-linearly with the depth. \textbf{Second}, the increasing depth introduces optimization difficulty~\cite{he2016deep}. Although ResNets seem to overcome the dilemma, managing to train a network with hundreds of layers, some works~\cite{veit2016residual,de2020batch} indicate ResNets might not be as deep as they appear to be. For example, \cite{veit2016residual} suggests ResNets behave like ensembles of shallow networks, which implies the ERFs of ResNets could still be very limited even if the depth dramatically increases. Such phenomenon is also empirically observed in previous works~\cite{kim2021dead}. To summarize, large kernels design requires fewer layers to obtain large ERFs and avoids the optimization issue brought by the increasing depth. 

	To support our viewpoint, we choose ResNet-101/152 and the aforementioned RepLKNet-13/31 as the representatives of small-kernel and large-kernel models, which are all well-trained on ImageNet, and test with 50 images from the ImageNet validation set resized to 1024$\times$1024. To visualize the ERF, we use a simple yet effective method (code released at \cite{RepLKNet-pytorch}) as introduced in Appendix B, following \cite{kim2021dead}. Briefly, we produce an \emph{aggregated contribution score matrix} $\mathrm{A}$ (1024$\times$1024), where each entry $a$ ($0\leq a \leq 1$) measures the contribution of the corresponding pixel on the input image to the central point of the feature map produced by the last layer. Fig.~\ref{fig-rf} shows the high-contribution pixels of ResNet-101 gather around the central point, but the outer points have very low contributions, indicating a limited ERF. ResNet-152 shows a similar pattern, suggesting the more 3$\times$3 layers do not significantly increase the ERF. On the other hand, the high-contribution pixels in Fig.~\ref{fig-rf} (C) are more evenly distributed, suggesting RepLKNet-13 attends to more outer pixels. With larger kernels, RepLKNet-31 makes the high-contribution pixels spread more uniformly, indicating an even larger ERF. 
	
		\setlength{\tabcolsep}{4pt}
	\begin{table}[t]
		\caption{Quantitative analysis on the ERF with the high-contribution area ratio $r$. A larger $r$ suggests a smoother distribution of high-contribution pixels, hence larger ERF.}
		\label{table-rf}
		\vspace{-0.2in}
		\begin{center}
			\small
			\begin{tabular}{lcccc}
				\hline
				& $t=20\%$		&	$t=30\%$			&	$t=50\%$	&	$t=99\%$\\
				\hline
				ResNet-101		&	0.9\%		&	1.5\%	&	3.2\%	& 22.4\%	\\
				ResNet-152		&	1.1\%		&	1.8\%	&	3.9\%	& 34.4\%\\
				RepLKNet-13		&	11.2\%		&	17.1\%	&	30.2\%	&	96.3\%\\		
				RepLKNet-31		&	16.3\%		&	24.7\%	&   43.2\%	&	98.6\%	\\
				\hline
			\end{tabular}
			
		\end{center}
		\vspace{-0.2in}
	\end{table}
	
	Table~\ref{table-rf} presents a quantitative analysis, where we report the high-contribution area ratio $r$ of a minimum rectangle that covers the contribution scores over a given threshold $t$. For examples, 20\% of the pixel contributions ($\mathrm{A}$ values) of ResNet-101 reside within a 103$\times$103 area at the center, so that the area ratio is $(103/1024)^2=1.0\%$ with $t=20\%$. We make several intriguing observations. \textbf{1)} While being significantly deeper, ResNets have much smaller ERFs than RepLKNets. For example, over 99\% of the contribution scores of ResNet-101 reside within a small area which takes up only 23.4\% of the total area, while such area ratio of RepLKNet-31 is 98.6\%, which means most of pixels considerably contribute to the final predictions. \textbf{2)} Adding more layers to ResNet-101 does not effectively enlarge the ERF, while scaling up the kernels improves the ERF with marginal computational costs.

	\subsection{Large-Kernel Models are More Similar to Human in Shape Bias}
	
	We have found out that RepLKNet-31B has much higher shape bias than Swin Transformer and small-kernel CNNs.
	
	\begin{figure}[t]
		\begin{center}
			\includegraphics[width=\linewidth]{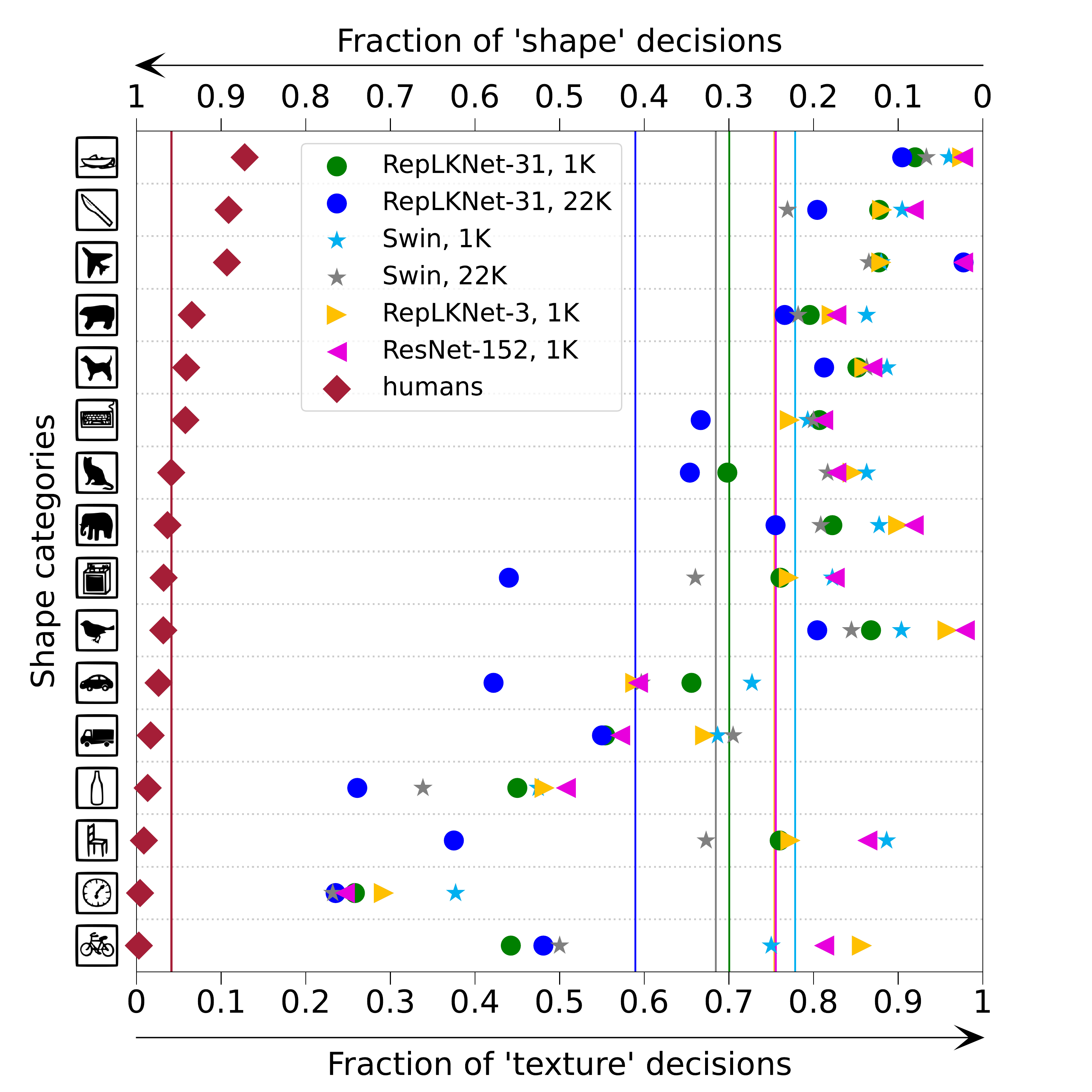}
			\vspace{-0.25in}
			\caption{Shape bias of RepLKNet, Swin, and ResNet-152 pretrained on ImageNet-1K or 22K. The scatters represent the shape bias of 16 categories, and the vertical lines are the averages across categories (note RepLKNet-3 and ResNet-152 are very close).}
			\label{fig-shapebias}
			\vspace{-0.25in}
		\end{center}
	\end{figure}
	
	A recent work~\cite{tuli2021convolutional} reported that vision transformers are more similar to the human vision systems in that they make predictions more based on the overall shapes of objects, while CNNs focus more on the local textures. We follow its methodology and use its toolbox~\cite{modelvshuman} to obtain the shape bias (\eg, the fraction of predictions made based on the shapes, rather than the textures) of RepLKNet-31B and Swin-B pretrained on ImageNet-1K or 22K, together with two small-kernel baselines, RepLKNet-3 and ResNet-152. Fig.~\ref{fig-shapebias} shows that RepLKNet has higher shape bias than Swin. Considering RepLKNet and Swin have similar overall architectures, we reckon shape bias is closely related to the \emph{Effective Receptive Field} rather than the concrete formulation of self-attention (\ie, the \emph{query-key-value} design). This also explains \textbf{1)} the high shape bias of \emph{ViTs}~\cite{vit} reported by \cite{tuli2021convolutional} (since ViTs employ global attention), \textbf{2)} the low shape bias of 1K-pretrained Swin (attention within local windows), and \textbf{3)} the shape bias of the small-kernel baseline RepLKNet-3, which is very close to ResNet-152 (both models are composed of $3\times 3$ convolutions).

	\subsection{Large Kernel Design is a Generic Design Element that Works with ConvNeXt} Replacing the 7$\times$7 convolutions in ConvNeXt~\cite{liu2022convnet} by kernels as large as 31$\times$31 brings significant improvements, \eg, ConNeXt-Tiny + large kernel \textgreater ConNeXt-Small , and ConNeXt-Small + large kernel \textgreater ConNeXt-Base. 
	
		\setlength{\tabcolsep}{3pt}
	\begin{table*}
		\caption{ConvNeXt with different kernel sizes. The models are pretrained on ImageNet-1K in 120 epochs with 224$\times$224 input and finetuned on ADE20K with UperNet in 80K iterations. On ADE20K, we test the \emph{single-scale} mIoU, and compute the FLOPs with input of 2048$\times$512, following Swin.}
		\label{table-convnext}
		\vspace{-0.25in}
		\begin{center}		
			\small
			\begin{tabular}{ll|ccc|ccc}
				\hline
				& & \multicolumn{3}{c|}{ImageNet} & \multicolumn{3}{c}{ADE20K} \\
				Kernel size 	&		Architecture		&	Top-1		& \makecell{Params}		&\makecell{FLOPs}	&	mIoU	& Params	& FLOPs	\\
				\hline
				7-7-7-7			&	ConvNeXt-Tiny	&	     81.0           &   29M       &   4.5G   &   44.6  	 	&   60M		&   939G\\
				7-7-7-7			&	ConvNeXt-Small	&	     82.1           &   50M       &   8.7G   &   45.9  	 	&   82M  	&   1027G\\
				7-7-7-7			&	ConvNeXt-Base	&	     82.8           &   89M       &   15.4G  &   47.2  	 	&   122M  	&   1170G\\
				\hline
				31-29-27-13		&	ConvNeXt-Tiny	&	     81.6           &   32M       &   6.1G   &   \textbf{46.2}  	 	&   64M  &   973G\\
				31-29-27-13		&	ConvNeXt-Small	&	     82.5           &   58M       &   11.3G  &   \textbf{48.2}  	 	&   90M  &   1081G\\
				\hline
			\end{tabular}
		\end{center}
		\vspace{-0.3in}
	\end{table*}
	
	We use the recently proposed ConvNeXt~\cite{liu2022convnet} as the benchmark architecture to evaluate large kernel as a \emph{generic design element}. We simply replace the 7$\times$7 convolutions in ConvNeXt~\cite{liu2022convnet} by kernels as large as 31$\times$31. The training configurations on ImageNet (120 epochs) and ADE20K (80K iterations) are identical to the results shown in Sec.~\ref{sect-replknet-120}. Table~\ref{table-convnext} shows that though the original kernels are already 7$\times$7, further increasing the kernel sizes still brings significant improvements, especially on the downstream task: with kernels as large as 31$\times$31, ConvNeXt-Tiny outperforms the original ConvNeXt-Small, and the large-kernel ConvNeXt-Small outperforms the original ConvNeXt-Base. Again, such phenomena demonstrate that kernel size is an important scaling dimension.

	\subsection{Large Kernels Outperform Small Kernels with High Dilation Rates} 
	
	Please refer to Appendix C for details.

	\section{Limitations}
	
	Although large kernel design greatly improves CNNs on both ImageNet and downstream tasks, however, according to Table~\ref{table-to-swin}, as the scale of data and model increases, \emph{RepLKNets} start to fall behind \emph{Swin Transformers}, \eg, the ImageNet top-1 accuracy of RepLKNet-31L is 0.7\% lower than Swin-L with ImageNet-22K pretraining (while the downstream scores are still comparable). It is not clear whether the gap is resulted from suboptimal hyper-parameter tuning or some other fundamental drawback of CNNs which emerges when data/model scales up. We are working in progress on the problem. 
	
	\section{Conclusion}
	
	This paper revisits large convolutional kernels, which have long been neglected in designing \emph{CNN} architectures. We demonstrate that using a few large kernels instead of many small kernels results in larger \emph{effective receptive field} more efficiently, boosting CNN's performances especially on downstream tasks by a large margin, and greatly closing the performance gap between CNNs and ViTs when data and models scale up. We hope our work could advance both studies of CNNs and ViTs. On one hand, for CNN community, our findings suggest that we should pay special attention to \emph{ERFs}, which may be the key to high performances. On the other hand, for ViT community, since large convolutions act as an alternative to \emph{multi-head self-attentions} with similar behaviors, it may help to understand the intrinsic mechanism of self-attentions.

	{\small
		\bibliographystyle{ieee_fullname}
		\bibliography{replknetbib}

\begin{thebibliography}{100}\itemsep=-1pt

\bibitem{MegEngine}
Megengine:a fast, scalable and easy-to-use deep learning framework.
\newblock \url{https://github.com/MegEngine/MegEngine}, 2020.

\bibitem{RepLKNet-pytorch}
Official pytorch implementation of replknet.
\newblock \url{https://github.com/DingXiaoH/RepLKNet-pytorch}, 2022.

\bibitem{ba2016layer}
Jimmy~Lei Ba, Jamie~Ryan Kiros, and Geoffrey~E Hinton.
\newblock Layer normalization.
\newblock {\em arXiv preprint arXiv:1607.06450}, 2016.

\bibitem{bao2021beit}
Hangbo Bao, Li Dong, and Furu Wei.
\newblock Beit: Bert pre-training of image transformers.
\newblock {\em arXiv preprint arXiv:2106.08254}, 2021.

\bibitem{bello2019attention}
Irwan Bello, Barret Zoph, Ashish Vaswani, Jonathon Shlens, and Quoc~V Le.
\newblock Attention augmented convolutional networks.
\newblock In {\em Proceedings of the IEEE/CVF international conference on
  computer vision}, pages 3286--3295, 2019.

\bibitem{modelvshuman}
bethgelab.
\newblock Toolbox of model-vs-human.
\newblock \url{https://github.com/bethgelab/model-vs-human}, 2022.

\bibitem{bodla2017soft}
Navaneeth Bodla, Bharat Singh, Rama Chellappa, and Larry~S Davis.
\newblock Soft-nms--improving object detection with one line of code.
\newblock In {\em Proceedings of the IEEE international conference on computer
  vision}, pages 5561--5569, 2017.

\bibitem{brendel2019approximating}
Wieland Brendel and Matthias Bethge.
\newblock Approximating cnns with bag-of-local-features models works
  surprisingly well on imagenet.
\newblock {\em arXiv preprint arXiv:1904.00760}, 2019.

\bibitem{cai2019cascade}
Zhaowei Cai and Nuno Vasconcelos.
\newblock Cascade r-cnn: High quality object detection and instance
  segmentation.
\newblock {\em IEEE Transactions on Pattern Analysis and Machine Intelligence},
  2019.

\bibitem{dino}
Mathilde Caron, Hugo Touvron, Ishan Misra, Herv{\'e} J{\'e}gou, Julien Mairal,
  Piotr Bojanowski, and Armand Joulin.
\newblock Emerging properties in self-supervised vision transformers.
\newblock {\em arXiv preprint arXiv:2104.14294}, 2021.

\bibitem{ipt}
Hanting Chen, Yunhe Wang, Tianyu Guo, Chang Xu, Yiping Deng, Zhenhua Liu, Siwei
  Ma, Chunjing Xu, Chao Xu, and Wen Gao.
\newblock Pre-trained image processing transformer.
\newblock In {\em Proceedings of the IEEE/CVF Conference on Computer Vision and
  Pattern Recognition}, pages 12299--12310, 2021.

\bibitem{chen2019hybrid}
Kai Chen, Jiangmiao Pang, Jiaqi Wang, Yu Xiong, Xiaoxiao Li, Shuyang Sun,
  Wansen Feng, Ziwei Liu, Jianping Shi, Wanli Ouyang, et~al.
\newblock Hybrid task cascade for instance segmentation.
\newblock In {\em Proceedings of the IEEE/CVF Conference on Computer Vision and
  Pattern Recognition}, pages 4974--4983, 2019.

\bibitem{mmdetection}
Kai Chen, Jiaqi Wang, Jiangmiao Pang, Yuhang Cao, Yu Xiong, Xiaoxiao Li,
  Shuyang Sun, Wansen Feng, Ziwei Liu, Jiarui Xu, Zheng Zhang, Dazhi Cheng,
  Chenchen Zhu, Tianheng Cheng, Qijie Zhao, Buyu Li, Xin Lu, Rui Zhu, Yue Wu,
  Jifeng Dai, Jingdong Wang, Jianping Shi, Wanli Ouyang, Chen~Change Loy, and
  Dahua Lin.
\newblock {MMDetection}: Open mmlab detection toolbox and benchmark.
\newblock {\em arXiv preprint arXiv:1906.07155}, 2019.

\bibitem{chen2017deeplab}
Liang-Chieh Chen, George Papandreou, Iasonas Kokkinos, Kevin Murphy, and Alan~L
  Yuille.
\newblock Deeplab: Semantic image segmentation with deep convolutional nets,
  atrous convolution, and fully connected crfs.
\newblock {\em IEEE transactions on pattern analysis and machine intelligence},
  40(4):834--848, 2017.

\bibitem{chen2017rethinking}
Liang-Chieh Chen, George Papandreou, Florian Schroff, and Hartwig Adam.
\newblock Rethinking atrous convolution for semantic image segmentation.
\newblock {\em arXiv preprint arXiv:1706.05587}, 2017.

\bibitem{chen2018encoder}
Liang-Chieh Chen, Yukun Zhu, George Papandreou, Florian Schroff, and Hartwig
  Adam.
\newblock Encoder-decoder with atrous separable convolution for semantic image
  segmentation.
\newblock In {\em Proceedings of the European conference on computer vision
  (ECCV)}, pages 801--818, 2018.

\bibitem{mocov3}
Xinlei Chen, Saining Xie, and Kaiming He.
\newblock An empirical study of training self-supervised visual transformers.
\newblock {\em arXiv e-prints}, pages arXiv--2104, 2021.

\bibitem{chollet2017xception}
Fran{\c{c}}ois Chollet.
\newblock Xception: Deep learning with depthwise separable convolutions.
\newblock In {\em Proceedings of the IEEE conference on computer vision and
  pattern recognition}, pages 1251--1258, 2017.

\bibitem{chu2021conditional}
Xiangxiang Chu, Zhi Tian, Bo Zhang, Xinlong Wang, Xiaolin Wei, Huaxia Xia, and
  Chunhua Shen.
\newblock Conditional positional encodings for vision transformers.
\newblock {\em arXiv preprint arXiv:2102.10882}, 2021.

\bibitem{mmseg2020}
MMSegmentation Contributors.
\newblock {MMSegmentation}: Openmmlab semantic segmentation toolbox and
  benchmark.
\newblock \url{https://github.com/open-mmlab/mmsegmentation}, 2020.

\bibitem{cordonnier2019relationship}
Jean-Baptiste Cordonnier, Andreas Loukas, and Martin Jaggi.
\newblock On the relationship between self-attention and convolutional layers.
\newblock {\em arXiv preprint arXiv:1911.03584}, 2019.

\bibitem{cityscapes}
Marius Cordts, Mohamed Omran, Sebastian Ramos, Timo Rehfeld, Markus Enzweiler,
  Rodrigo Benenson, Uwe Franke, Stefan Roth, and Bernt Schiele.
\newblock The cityscapes dataset for semantic urban scene understanding.
\newblock In {\em 2016 {IEEE} Conference on Computer Vision and Pattern
  Recognition, {CVPR} 2016, Las Vegas, NV, USA, June 27-30, 2016}, pages
  3213--3223. {IEEE} Computer Society, 2016.

\bibitem{cubuk2020randaugment}
Ekin~D Cubuk, Barret Zoph, Jonathon Shlens, and Quoc~V Le.
\newblock Randaugment: Practical automated data augmentation with a reduced
  search space.
\newblock In {\em Proceedings of the IEEE/CVF Conference on Computer Vision and
  Pattern Recognition Workshops}, pages 702--703, 2020.

\bibitem{dai2017deformable}
Jifeng Dai, Haozhi Qi, Yuwen Xiong, Yi Li, Guodong Zhang, Han Hu, and Yichen
  Wei.
\newblock Deformable convolutional networks.
\newblock In {\em Proceedings of the IEEE international conference on computer
  vision}, pages 764--773, 2017.

\bibitem{dai2021dynamic}
Xiyang Dai, Yinpeng Chen, Bin Xiao, Dongdong Chen, Mengchen Liu, Lu Yuan, and
  Lei Zhang.
\newblock Dynamic head: Unifying object detection heads with attentions.
\newblock In {\em Proceedings of the IEEE/CVF Conference on Computer Vision and
  Pattern Recognition}, pages 7373--7382, 2021.

\bibitem{de2020batch}
Soham De and Samuel~L Smith.
\newblock Batch normalization biases residual blocks towards the identity
  function in deep networks.
\newblock {\em arXiv preprint arXiv:2002.10444}, 2020.

\bibitem{ding2021repmlpnet}
Xiaohan Ding, Honghao Chen, Xiangyu Zhang, Jungong Han, and Guiguang Ding.
\newblock Repmlpnet: Hierarchical vision mlp with re-parameterized locality.
\newblock {\em arXiv preprint arXiv:2112.11081}, 2021.

\bibitem{ding2019acnet}
Xiaohan Ding, Yuchen Guo, Guiguang Ding, and Jungong Han.
\newblock Acnet: Strengthening the kernel skeletons for powerful cnn via
  asymmetric convolution blocks.
\newblock In {\em Proceedings of the IEEE International Conference on Computer
  Vision}, pages 1911--1920, 2019.

\bibitem{ding2021resrep}
Xiaohan Ding, Tianxiang Hao, Jianchao Tan, Ji Liu, Jungong Han, Yuchen Guo, and
  Guiguang Ding.
\newblock Resrep: Lossless cnn pruning via decoupling remembering and
  forgetting.
\newblock In {\em Proceedings of the IEEE/CVF International Conference on
  Computer Vision}, pages 4510--4520, 2021.

\bibitem{ding2021diverse}
Xiaohan Ding, Xiangyu Zhang, Jungong Han, and Guiguang Ding.
\newblock Diverse branch block: Building a convolution as an inception-like
  unit.
\newblock In {\em Proceedings of the IEEE/CVF Conference on Computer Vision and
  Pattern Recognition}, pages 10886--10895, 2021.

\bibitem{ding2021repvgg}
Xiaohan Ding, Xiangyu Zhang, Ningning Ma, Jungong Han, Guiguang Ding, and Jian
  Sun.
\newblock Repvgg: Making vgg-style convnets great again.
\newblock In {\em Proceedings of the IEEE/CVF Conference on Computer Vision and
  Pattern Recognition}, pages 13733--13742, 2021.

\bibitem{dollar2021fast}
Piotr Doll{\'a}r, Mannat Singh, and Ross Girshick.
\newblock Fast and accurate model scaling.
\newblock In {\em Proceedings of the IEEE/CVF Conference on Computer Vision and
  Pattern Recognition}, pages 924--932, 2021.

\bibitem{dong2021cswin}
Xiaoyi Dong, Jianmin Bao, Dongdong Chen, Weiming Zhang, Nenghai Yu, Lu Yuan,
  Dong Chen, and Baining Guo.
\newblock Cswin transformer: A general vision transformer backbone with
  cross-shaped windows.
\newblock {\em arXiv preprint arXiv:2107.00652}, 2021.

\bibitem{dong2021attention}
Yihe Dong, Jean-Baptiste Cordonnier, and Andreas Loukas.
\newblock Attention is not all you need: Pure attention loses rank doubly
  exponentially with depth.
\newblock {\em arXiv preprint arXiv:2103.03404}, 2021.

\bibitem{vit}
Alexey Dosovitskiy, Lucas Beyer, Alexander Kolesnikov, Dirk Weissenborn,
  Xiaohua Zhai, Thomas Unterthiner, Mostafa Dehghani, Matthias Minderer, Georg
  Heigold, Sylvain Gelly, Jakob Uszkoreit, and Neil Houlsby.
\newblock An image is worth 16x16 words: Transformers for image recognition at
  scale.
\newblock In {\em 9th International Conference on Learning Representations,
  {ICLR} 2021, Virtual Event, Austria, May 3-7, 2021}. OpenReview.net, 2021.

\bibitem{geirhos2018imagenet}
Robert Geirhos, Patricia Rubisch, Claudio Michaelis, Matthias Bethge, Felix~A
  Wichmann, and Wieland Brendel.
\newblock Imagenet-trained cnns are biased towards texture; increasing shape
  bias improves accuracy and robustness.
\newblock {\em arXiv preprint arXiv:1811.12231}, 2018.

\bibitem{ghiasi2021multi}
Golnaz Ghiasi, Barret Zoph, Ekin~D Cubuk, Quoc~V Le, and Tsung-Yi Lin.
\newblock Multi-task self-training for learning general representations.
\newblock In {\em Proceedings of the IEEE/CVF International Conference on
  Computer Vision}, pages 8856--8865, 2021.

\bibitem{gong2021improve}
Chengyue Gong, Dilin Wang, Meng Li, Vikas Chandra, and Qiang Liu.
\newblock Improve vision transformers training by suppressing over-smoothing.
\newblock {\em arXiv preprint arXiv:2104.12753}, 2021.

\bibitem{guo2020single}
Zichao Guo, Xiangyu Zhang, Haoyuan Mu, Wen Heng, Zechun Liu, Yichen Wei, and
  Jian Sun.
\newblock Single path one-shot neural architecture search with uniform
  sampling.
\newblock In {\em European Conference on Computer Vision}, pages 544--560.
  Springer, 2020.

\bibitem{han2021demystifying}
Qi Han, Zejia Fan, Qi Dai, Lei Sun, Ming-Ming Cheng, Jiaying Liu, and Jingdong
  Wang.
\newblock Demystifying local vision transformer: Sparse connectivity, weight
  sharing, and dynamic weight.
\newblock {\em arXiv preprint arXiv:2106.04263}, 2021.

\bibitem{he2017mask}
Kaiming He, Georgia Gkioxari, Piotr Doll{\'a}r, and Ross Girshick.
\newblock Mask r-cnn.
\newblock In {\em Proceedings of the IEEE international conference on computer
  vision}, pages 2961--2969, 2017.

\bibitem{he2016deep}
Kaiming He, Xiangyu Zhang, Shaoqing Ren, and Jian Sun.
\newblock Deep residual learning for image recognition.
\newblock In {\em Proceedings of the IEEE conference on computer vision and
  pattern recognition}, pages 770--778, 2016.

\bibitem{hendrycks2016gaussian}
Dan Hendrycks and Kevin Gimpel.
\newblock Gaussian error linear units (gelus).
\newblock {\em arXiv preprint arXiv:1606.08415}, 2016.

\bibitem{hinton2021represent}
Geoffrey Hinton.
\newblock How to represent part-whole hierarchies in a neural network.
\newblock {\em arXiv preprint arXiv:2102.12627}, 2021.

\bibitem{mbv3}
Andrew Howard, Ruoming Pang, Hartwig Adam, Quoc~V. Le, Mark Sandler, Bo Chen,
  Weijun Wang, Liang{-}Chieh Chen, Mingxing Tan, Grace Chu, Vijay Vasudevan,
  and Yukun Zhu.
\newblock Searching for mobilenetv3.
\newblock In {\em 2019 {IEEE/CVF} International Conference on Computer Vision,
  {ICCV} 2019, Seoul, Korea (South), October 27 - November 2, 2019}, pages
  1314--1324. {IEEE}, 2019.

\bibitem{mbv1}
Andrew~G Howard, Menglong Zhu, Bo Chen, Dmitry Kalenichenko, Weijun Wang,
  Tobias Weyand, Marco Andreetto, and Hartwig Adam.
\newblock Mobilenets: Efficient convolutional neural networks for mobile vision
  applications.
\newblock {\em arXiv preprint arXiv:1704.04861}, 2017.

\bibitem{hu2019local}
Han Hu, Zheng Zhang, Zhenda Xie, and Stephen Lin.
\newblock Local relation networks for image recognition.
\newblock In {\em Proceedings of the IEEE/CVF International Conference on
  Computer Vision}, pages 3464--3473, 2019.

\bibitem{hu2018squeeze}
Jie Hu, Li Shen, and Gang Sun.
\newblock Squeeze-and-excitation networks.
\newblock In {\em Proceedings of the IEEE conference on computer vision and
  pattern recognition}, pages 7132--7141, 2018.

\bibitem{huang2017densely}
Gao Huang, Zhuang Liu, Laurens van~der Maaten, and Kilian~Q. Weinberger.
\newblock Densely connected convolutional networks.
\newblock In {\em 2017 {IEEE} Conference on Computer Vision and Pattern
  Recognition, {CVPR} 2017, Honolulu, HI, USA, July 21-26, 2017}, pages
  2261--2269. {IEEE} Computer Society, 2017.

\bibitem{stochastic}
Gao Huang, Yu Sun, Zhuang Liu, Daniel Sedra, and Kilian~Q. Weinberger.
\newblock Deep networks with stochastic depth.
\newblock In Bastian Leibe, Jiri Matas, Nicu Sebe, and Max Welling, editors,
  {\em Computer Vision - {ECCV} 2016 - 14th European Conference, Amsterdam, The
  Netherlands, October 11-14, 2016, Proceedings, Part {IV}}, volume 9908 of
  {\em Lecture Notes in Computer Science}, pages 646--661. Springer, 2016.

\bibitem{ioffe2015batch}
Sergey Ioffe and Christian Szegedy.
\newblock Batch normalization: Accelerating deep network training by reducing
  internal covariate shift.
\newblock In {\em International Conference on Machine Learning}, pages
  448--456, 2015.

\bibitem{clip}
Chao Jia, Yinfei Yang, Ye Xia, Yi-Ting Chen, Zarana Parekh, Hieu Pham, Quoc~V
  Le, Yunhsuan Sung, Zhen Li, and Tom Duerig.
\newblock Scaling up visual and vision-language representation learning with
  noisy text supervision.
\newblock {\em arXiv preprint arXiv:2102.05918}, 2021.

\bibitem{kayhan2020translation}
Osman~Semih Kayhan and Jan C~van Gemert.
\newblock On translation invariance in cnns: Convolutional layers can exploit
  absolute spatial location.
\newblock In {\em Proceedings of the IEEE/CVF Conference on Computer Vision and
  Pattern Recognition}, pages 14274--14285, 2020.

\bibitem{kim2021dead}
Bum~Jun Kim, Hyeyeon Choi, Hyeonah Jang, Dong~Gu Lee, Wonseok Jeong, and
  Sang~Woo Kim.
\newblock Dead pixel test using effective receptive field.
\newblock {\em arXiv preprint arXiv:2108.13576}, 2021.

\bibitem{krizhevsky2012imagenet}
Alex Krizhevsky, Ilya Sutskever, and Geoffrey~E Hinton.
\newblock Imagenet classification with deep convolutional neural networks.
\newblock In {\em Advances in neural information processing systems}, pages
  1097--1105, 2012.

\bibitem{liang2021swinir}
Jingyun Liang, Jiezhang Cao, Guolei Sun, Kai Zhang, Luc Van~Gool, and Radu
  Timofte.
\newblock Swinir: Image restoration using swin transformer.
\newblock In {\em Proceedings of the IEEE/CVF International Conference on
  Computer Vision}, pages 1833--1844, 2021.

\bibitem{lin2014microsoft}
Tsung-Yi Lin, Michael Maire, Serge Belongie, James Hays, Pietro Perona, Deva
  Ramanan, Piotr Doll{\'a}r, and C~Lawrence Zitnick.
\newblock Microsoft coco: Common objects in context.
\newblock In {\em European conference on computer vision}, pages 740--755.
  Springer, 2014.

\bibitem{liu2018darts}
Hanxiao Liu, Karen Simonyan, and Yiming Yang.
\newblock Darts: Differentiable architecture search.
\newblock {\em arXiv preprint arXiv:1806.09055}, 2018.

\bibitem{liu2020understanding}
Liyuan Liu, Xiaodong Liu, Jianfeng Gao, Weizhu Chen, and Jiawei Han.
\newblock Understanding the difficulty of training transformers.
\newblock {\em arXiv preprint arXiv:2004.08249}, 2020.

\bibitem{liu2021swin}
Ze Liu, Han Hu, Yutong Lin, Zhuliang Yao, Zhenda Xie, Yixuan Wei, Jia Ning, Yue
  Cao, Zheng Zhang, Li Dong, et~al.
\newblock Swin transformer v2: Scaling up capacity and resolution.
\newblock {\em arXiv preprint arXiv:2111.09883}, 2021.

\bibitem{swin}
Ze Liu, Yutong Lin, Yue Cao, Han Hu, Yixuan Wei, Zheng Zhang, Stephen Lin, and
  Baining Guo.
\newblock Swin transformer: Hierarchical vision transformer using shifted
  windows.
\newblock In {\em Proceedings of the IEEE/CVF International Conference on
  Computer Vision}, pages 10012--10022, 2021.

\bibitem{liu2022convnet}
Zhuang Liu, Hanzi Mao, Chao-Yuan Wu, Christoph Feichtenhofer, Trevor Darrell,
  and Saining Xie.
\newblock A convnet for the 2020s.
\newblock {\em arXiv preprint arXiv:2201.03545}, 2022.

\bibitem{long2015fully}
Jonathan Long, Evan Shelhamer, and Trevor Darrell.
\newblock Fully convolutional networks for semantic segmentation.
\newblock In {\em Proceedings of the IEEE Conference on Computer Vision and
  Pattern Recognition}, pages 3431--3440, 2015.

\bibitem{loshchilov2017decoupled}
Ilya Loshchilov and Frank Hutter.
\newblock Decoupled weight decay regularization.
\newblock {\em arXiv preprint arXiv:1711.05101}, 2017.

\bibitem{erf}
Wenjie Luo, Yujia Li, Raquel Urtasun, and Richard~S. Zemel.
\newblock Understanding the effective receptive field in deep convolutional
  neural networks.
\newblock In Daniel~D. Lee, Masashi Sugiyama, Ulrike von Luxburg, Isabelle
  Guyon, and Roman Garnett, editors, {\em Advances in Neural Information
  Processing Systems 29: Annual Conference on Neural Information Processing
  Systems 2016, December 5-10, 2016, Barcelona, Spain}, pages 4898--4906, 2016.

\bibitem{ma2018shufflenet}
Ningning Ma, Xiangyu Zhang, Hai-Tao Zheng, and Jian Sun.
\newblock Shufflenet v2: Practical guidelines for efficient cnn architecture
  design.
\newblock In {\em Proceedings of the European conference on computer vision
  (ECCV)}, pages 116--131, 2018.

\bibitem{mathieu2013fast}
Micha{\"{e}}l Mathieu, Mikael Henaff, and Yann LeCun.
\newblock Fast training of convolutional networks through ffts.
\newblock In Yoshua Bengio and Yann LeCun, editors, {\em 2nd International
  Conference on Learning Representations, {ICLR} 2014, Banff, AB, Canada, April
  14-16, 2014, Conference Track Proceedings}, 2014.

\bibitem{paul2021vision}
Sayak Paul and Pin-Yu Chen.
\newblock Vision transformers are robust learners.
\newblock {\em arXiv preprint arXiv:2105.07581}, 2021.

\bibitem{peng2017large}
Chao Peng, Xiangyu Zhang, Gang Yu, Guiming Luo, and Jian Sun.
\newblock Large kernel matters--improve semantic segmentation by global
  convolutional network.
\newblock In {\em Proceedings of the IEEE conference on computer vision and
  pattern recognition}, pages 4353--4361, 2017.

\bibitem{regnet}
Ilija Radosavovic, Raj~Prateek Kosaraju, Ross Girshick, Kaiming He, and Piotr
  Doll{\'a}r.
\newblock Designing network design spaces.
\newblock In {\em Proceedings of the IEEE/CVF Conference on Computer Vision and
  Pattern Recognition}, pages 10428--10436, 2020.

\bibitem{raghu2021vision}
Maithra Raghu, Thomas Unterthiner, Simon Kornblith, Chiyuan Zhang, and Alexey
  Dosovitskiy.
\newblock Do vision transformers see like convolutional neural networks?
\newblock {\em arXiv preprint arXiv:2108.08810}, 2021.

\bibitem{sasa}
Prajit Ramachandran, Niki Parmar, Ashish Vaswani, Irwan Bello, Anselm Levskaya,
  and Jonathon Shlens.
\newblock Stand-alone self-attention in vision models.
\newblock {\em arXiv preprint arXiv:1906.05909}, 2019.

\bibitem{ranftl2021vision}
Ren{\'e} Ranftl, Alexey Bochkovskiy, and Vladlen Koltun.
\newblock Vision transformers for dense prediction.
\newblock In {\em Proceedings of the IEEE/CVF International Conference on
  Computer Vision}, pages 12179--12188, 2021.

\bibitem{rao2021global}
Yongming Rao, Wenliang Zhao, Zheng Zhu, Jiwen Lu, and Jie Zhou.
\newblock Global filter networks for image classification.
\newblock {\em arXiv preprint arXiv:2107.00645}, 2021.

\bibitem{romero2021flexconv}
David~W Romero, Robert-Jan Bruintjes, Jakub~M Tomczak, Erik~J Bekkers, Mark
  Hoogendoorn, and Jan~C van Gemert.
\newblock Flexconv: Continuous kernel convolutions with differentiable kernel
  sizes.
\newblock {\em arXiv preprint arXiv:2110.08059}, 2021.

\bibitem{romero2021ckconv}
David~W Romero, Anna Kuzina, Erik~J Bekkers, Jakub~M Tomczak, and Mark
  Hoogendoorn.
\newblock Ckconv: Continuous kernel convolution for sequential data.
\newblock {\em arXiv preprint arXiv:2102.02611}, 2021.

\bibitem{mbv2}
Mark Sandler, Andrew Howard, Menglong Zhu, Andrey Zhmoginov, and Liang-Chieh
  Chen.
\newblock Mobilenetv2: Inverted residuals and linear bottlenecks.
\newblock In {\em Proceedings of the IEEE conference on computer vision and
  pattern recognition}, pages 4510--4520, 2018.

\bibitem{shaw2018self}
Peter Shaw, Jakob Uszkoreit, and Ashish Vaswani.
\newblock Self-attention with relative position representations.
\newblock {\em arXiv preprint arXiv:1803.02155}, 2018.

\bibitem{simonyan2014very}
Karen Simonyan and Andrew Zisserman.
\newblock Very deep convolutional networks for large-scale image recognition.
\newblock {\em arXiv preprint arXiv:1409.1556}, 2014.

\bibitem{bot}
Aravind Srinivas, Tsung-Yi Lin, Niki Parmar, Jonathon Shlens, Pieter Abbeel,
  and Ashish Vaswani.
\newblock Bottleneck transformers for visual recognition.
\newblock In {\em Proceedings of the IEEE/CVF Conference on Computer Vision and
  Pattern Recognition}, pages 16519--16529, 2021.

\bibitem{szegedy2017inception}
Christian Szegedy, Sergey Ioffe, Vincent Vanhoucke, and Alexander~A Alemi.
\newblock Inception-v4, inception-resnet and the impact of residual connections
  on learning.
\newblock In {\em Thirty-first AAAI conference on artificial intelligence},
  2017.

\bibitem{szegedy2015going}
Christian Szegedy, Wei Liu, Yangqing Jia, Pierre Sermanet, Scott Reed, Dragomir
  Anguelov, Dumitru Erhan, Vincent Vanhoucke, and Andrew Rabinovich.
\newblock Going deeper with convolutions.
\newblock In {\em Proceedings of the IEEE conference on computer vision and
  pattern recognition}, pages 1--9, 2015.

\bibitem{szegedy2016rethinking}
Christian Szegedy, Vincent Vanhoucke, Sergey Ioffe, Jon Shlens, and Zbigniew
  Wojna.
\newblock Rethinking the inception architecture for computer vision.
\newblock In {\em Proceedings of the IEEE conference on computer vision and
  pattern recognition}, pages 2818--2826, 2016.

\bibitem{efficientnet}
Mingxing Tan and Quoc~V Le.
\newblock Efficientnet: Rethinking model scaling for convolutional neural
  networks.
\newblock {\em arXiv preprint arXiv:1905.11946}, 2019.

\bibitem{thomee2016yfcc100m}
Bart Thomee, David~A Shamma, Gerald Friedland, Benjamin Elizalde, Karl Ni,
  Douglas Poland, Damian Borth, and Li-Jia Li.
\newblock Yfcc100m: The new data in multimedia research.
\newblock {\em Communications of the ACM}, 59(2):64--73, 2016.

\bibitem{fcos}
Zhi Tian, Chunhua Shen, Hao Chen, and Tong He.
\newblock {FCOS:} fully convolutional one-stage object detection.
\newblock In {\em 2019 {IEEE/CVF} International Conference on Computer Vision,
  {ICCV} 2019, Seoul, Korea (South), October 27 - November 2, 2019}, pages
  9626--9635. {IEEE}, 2019.

\bibitem{tolstikhin2021mlp}
Ilya Tolstikhin, Neil Houlsby, Alexander Kolesnikov, Lucas Beyer, Xiaohua Zhai,
  Thomas Unterthiner, Jessica Yung, Andreas Steiner, Daniel Keysers, Jakob
  Uszkoreit, et~al.
\newblock Mlp-mixer: An all-mlp architecture for vision.
\newblock {\em arXiv preprint arXiv:2105.01601}, 2021.

\bibitem{touvron2021resmlp}
Hugo Touvron, Piotr Bojanowski, Mathilde Caron, Matthieu Cord, Alaaeldin
  El-Nouby, Edouard Grave, Gautier Izacard, Armand Joulin, Gabriel Synnaeve,
  Jakob Verbeek, et~al.
\newblock Resmlp: Feedforward networks for image classification with
  data-efficient training.
\newblock {\em arXiv preprint arXiv:2105.03404}, 2021.

\bibitem{deit}
Hugo Touvron, Matthieu Cord, Matthijs Douze, Francisco Massa, Alexandre
  Sablayrolles, and Herv{\'e} J{\'e}gou.
\newblock Training data-efficient image transformers \& distillation through
  attention.
\newblock In {\em International Conference on Machine Learning}, pages
  10347--10357. PMLR, 2021.

\bibitem{trockman2022patches}
Asher Trockman and J~Zico Kolter.
\newblock Patches are all you need?
\newblock {\em arXiv preprint arXiv:2201.09792}, 2022.

\bibitem{tuli2021convolutional}
Shikhar Tuli, Ishita Dasgupta, Erin Grant, and Thomas~L Griffiths.
\newblock Are convolutional neural networks or transformers more like human
  vision?
\newblock {\em arXiv preprint arXiv:2105.07197}, 2021.

\bibitem{halonet}
Ashish Vaswani, Prajit Ramachandran, Aravind Srinivas, Niki Parmar, Blake
  Hechtman, and Jonathon Shlens.
\newblock Scaling local self-attention for parameter efficient visual
  backbones.
\newblock In {\em Proceedings of the IEEE/CVF Conference on Computer Vision and
  Pattern Recognition}, pages 12894--12904, 2021.

\bibitem{vaswani2017attention}
Ashish Vaswani, Noam Shazeer, Niki Parmar, Jakob Uszkoreit, Llion Jones,
  Aidan~N Gomez, {\L}ukasz Kaiser, and Illia Polosukhin.
\newblock Attention is all you need.
\newblock In {\em Advances in neural information processing systems}, pages
  5998--6008, 2017.

\bibitem{veit2016residual}
Andreas Veit, Michael~J Wilber, and Serge Belongie.
\newblock Residual networks behave like ensembles of relatively shallow
  networks.
\newblock In {\em Advances in neural information processing systems}, pages
  550--558, 2016.

\bibitem{wang2020axial}
Huiyu Wang, Yukun Zhu, Bradley Green, Hartwig Adam, Alan Yuille, and
  Liang-Chieh Chen.
\newblock Axial-deeplab: Stand-alone axial-attention for panoptic segmentation.
\newblock In {\em European Conference on Computer Vision}, pages 108--126.
  Springer, 2020.

\bibitem{wang2020deep}
Jingdong Wang, Ke Sun, Tianheng Cheng, Borui Jiang, Chaorui Deng, Yang Zhao,
  Dong Liu, Yadong Mu, Mingkui Tan, Xinggang Wang, et~al.
\newblock Deep high-resolution representation learning for visual recognition.
\newblock {\em IEEE transactions on pattern analysis and machine intelligence},
  2020.

\bibitem{wang2018understanding}
Panqu Wang, Pengfei Chen, Ye Yuan, Ding Liu, Zehua Huang, Xiaodi Hou, and
  Garrison Cottrell.
\newblock Understanding convolution for semantic segmentation.
\newblock In {\em 2018 IEEE winter conference on applications of computer
  vision (WACV)}, pages 1451--1460. Ieee, 2018.

\bibitem{pvt}
Wenhai Wang, Enze Xie, Xiang Li, Deng-Ping Fan, Kaitao Song, Ding Liang, Tong
  Lu, Ping Luo, and Ling Shao.
\newblock Pyramid vision transformer: A versatile backbone for dense prediction
  without convolutions.
\newblock {\em arXiv preprint arXiv:2102.12122}, 2021.

\bibitem{timm-aa}
Ross Wightman.
\newblock Timm implementation of randaugment.
\newblock
  \url{https://github.com/rwightman/pytorch-image-models/blob/master/timm/data/auto_augment.py},
  2022.

\bibitem{wu2019pay}
Felix Wu, Angela Fan, Alexei Baevski, Yann~N Dauphin, and Michael Auli.
\newblock Pay less attention with lightweight and dynamic convolutions.
\newblock {\em arXiv preprint arXiv:1901.10430}, 2019.

\bibitem{wu2021cvt}
Haiping Wu, Bin Xiao, Noel Codella, Mengchen Liu, Xiyang Dai, Lu Yuan, and Lei
  Zhang.
\newblock Cvt: Introducing convolutions to vision transformers.
\newblock In {\em Proceedings of the IEEE/CVF International Conference on
  Computer Vision}, pages 22--31, 2021.

\bibitem{xiao2018unified}
Tete Xiao, Yingcheng Liu, Bolei Zhou, Yuning Jiang, and Jian Sun.
\newblock Unified perceptual parsing for scene understanding.
\newblock In {\em Proceedings of the European Conference on Computer Vision
  (ECCV)}, pages 418--434, 2018.

\bibitem{xie2021segformer}
Enze Xie, Wenhai Wang, Zhiding Yu, Anima Anandkumar, Jose~M Alvarez, and Ping
  Luo.
\newblock Segformer: Simple and efficient design for semantic segmentation with
  transformers.
\newblock {\em arXiv preprint arXiv:2105.15203}, 2021.

\bibitem{xie2017aggregated}
Saining Xie, Ross Girshick, Piotr Doll{\'a}r, Zhuowen Tu, and Kaiming He.
\newblock Aggregated residual transformations for deep neural networks.
\newblock In {\em Proceedings of the IEEE conference on computer vision and
  pattern recognition}, pages 1492--1500, 2017.

\bibitem{swinself}
Zhenda Xie, Yutong Lin, Zhuliang Yao, Zheng Zhang, Qi Dai, Yue Cao, and Han Hu.
\newblock Self-supervised learning with swin transformers.
\newblock {\em arXiv preprint arXiv:2105.04553}, 2021.

\bibitem{yu2015multi}
Fisher Yu and Vladlen Koltun.
\newblock Multi-scale context aggregation by dilated convolutions.
\newblock {\em arXiv preprint arXiv:1511.07122}, 2015.

\bibitem{yu2017dilated}
Fisher Yu, Vladlen Koltun, and Thomas Funkhouser.
\newblock Dilated residual networks.
\newblock In {\em Proceedings of the IEEE conference on computer vision and
  pattern recognition}, pages 472--480, 2017.

\bibitem{yu2021metaformer}
Weihao Yu, Mi Luo, Pan Zhou, Chenyang Si, Yichen Zhou, Xinchao Wang, Jiashi
  Feng, and Shuicheng Yan.
\newblock Metaformer is actually what you need for vision.
\newblock {\em arXiv preprint arXiv:2111.11418}, 2021.

\bibitem{yuan2021volo}
Li Yuan, Qibin Hou, Zihang Jiang, Jiashi Feng, and Shuicheng Yan.
\newblock Volo: Vision outlooker for visual recognition.
\newblock {\em arXiv preprint arXiv:2106.13112}, 2021.

\bibitem{yun2019cutmix}
Sangdoo Yun, Dongyoon Han, Seong~Joon Oh, Sanghyuk Chun, Junsuk Choe, and
  Youngjoon Yoo.
\newblock Cutmix: Regularization strategy to train strong classifiers with
  localizable features.
\newblock In {\em Proceedings of the IEEE/CVF International Conference on
  Computer Vision}, pages 6023--6032, 2019.

\bibitem{zhang2017mixup}
Hongyi Zhang, Moustapha Cisse, Yann~N Dauphin, and David Lopez-Paz.
\newblock mixup: Beyond empirical risk minimization.
\newblock {\em arXiv preprint arXiv:1710.09412}, 2017.

\bibitem{resnest}
Hang Zhang, Chongruo Wu, Zhongyue Zhang, Yi Zhu, Zhi Zhang, Haibin Lin, Yue
  Sun, Tong He, Jonas Mueller, R Manmatha, et~al.
\newblock Resnest: Split-attention networks.
\newblock {\em arXiv preprint arXiv:2004.08955}, 2020.

\bibitem{zhang2021basisnet}
Mingda Zhang, Chun-Te Chu, Andrey Zhmoginov, Andrew Howard, Brendan Jou, Yukun
  Zhu, Li Zhang, Rebecca Hwa, and Adriana Kovashka.
\newblock Basisnet: Two-stage model synthesis for efficient inference.
\newblock In {\em Proceedings of the IEEE/CVF Conference on Computer Vision and
  Pattern Recognition}, pages 3081--3090, 2021.

\bibitem{zhang2018shufflenet}
Xiangyu Zhang, Xinyu Zhou, Mengxiao Lin, and Jian Sun.
\newblock Shufflenet: An extremely efficient convolutional neural network for
  mobile devices.
\newblock In {\em Proceedings of the IEEE conference on computer vision and
  pattern recognition}, pages 6848--6856, 2018.

\bibitem{zhang2021collaboration}
Yikang Zhang, Zhuo Chen, and Zhao Zhong.
\newblock Collaboration of experts: Achieving 80\% top-1 accuracy on imagenet
  with 100m flops.
\newblock {\em arXiv preprint arXiv:2107.03815}, 2021.

\bibitem{zhao2021battle}
Yucheng Zhao, Guangting Wang, Chuanxin Tang, Chong Luo, Wenjun Zeng, and
  Zheng-Jun Zha.
\newblock A battle of network structures: An empirical study of cnn,
  transformer, and mlp.
\newblock {\em arXiv preprint arXiv:2108.13002}, 2021.

\bibitem{zheng2021rethinking}
Sixiao Zheng, Jiachen Lu, Hengshuang Zhao, Xiatian Zhu, Zekun Luo, Yabiao Wang,
  Yanwei Fu, Jianfeng Feng, Tao Xiang, Philip~HS Torr, et~al.
\newblock Rethinking semantic segmentation from a sequence-to-sequence
  perspective with transformers.
\newblock In {\em Proceedings of the IEEE/CVF Conference on Computer Vision and
  Pattern Recognition}, pages 6881--6890, 2021.

\bibitem{zhong2020random}
Zhun Zhong, Liang Zheng, Guoliang Kang, Shaozi Li, and Yi Yang.
\newblock Random erasing data augmentation.
\newblock In {\em Proceedings of the AAAI conference on artificial
  intelligence}, volume~34, pages 13001--13008, 2020.

\bibitem{zhou2017places}
Bolei Zhou, Agata Lapedriza, Aditya Khosla, Aude Oliva, and Antonio Torralba.
\newblock Places: A 10 million image database for scene recognition.
\newblock {\em IEEE Transactions on Pattern Analysis and Machine Intelligence},
  2017.

\bibitem{zhou2019semantic}
Bolei Zhou, Hang Zhao, Xavier Puig, Tete Xiao, Sanja Fidler, Adela Barriuso,
  and Antonio Torralba.
\newblock Semantic understanding of scenes through the ade20k dataset.
\newblock {\em International Journal of Computer Vision}, 127(3):302--321,
  2019.

\bibitem{zhu2019empirical}
Xizhou Zhu, Dazhi Cheng, Zheng Zhang, Stephen Lin, and Jifeng Dai.
\newblock An empirical study of spatial attention mechanisms in deep networks.
\newblock In {\em Proceedings of the IEEE/CVF International Conference on
  Computer Vision}, pages 6688--6697, 2019.

\bibitem{zoph2016neural}
Barret Zoph and Quoc~V Le.
\newblock Neural architecture search with reinforcement learning.
\newblock {\em arXiv preprint arXiv:1611.01578}, 2016.

\end{thebibliography}
	}
	
	\newpage

	\section*{Appendix A: Training Configurations}
	
	\subsection*{ImageNet-1K}
	
	For training MobileNet V2 models (Sec.~\ref{sect-3}), we use 8 GPUs, an SGD optimizer with momentum of 0.9, a batch size of 32 per GPU, input resolution of 224$\times$224, weight decay of $4\times10^{-5}$, learning rate schedule with 5-epoch warmup, initial value of 0.1 and cosine annealing for 100 epochs. For the data augmentation, we only use random cropping and left-right flipping, as a common practice.
	
	For training RepLKNet models (Sec.~\ref{sect-replknet-120}),we use 32 GPUs and a batch size of 64 per GPU to train for 120 epochs. The optimizer is AdamW~\cite{loshchilov2017decoupled} with momentum of 0.9 and weight decay of 0.05. The learning rate setting includes an initial value of $4\times 10^{-3}$, cosine annealing and 10-epoch warm-up. For the data augmentation and regularization, we use RandAugment~\cite{cubuk2020randaugment} (``rand-m9-mstd0.5-inc1'' as implemented by timm~\cite{timm-aa}), label smoothing coefficient of 0.1, mixup~\cite{zhang2017mixup} with $\alpha=0.8$, CutMix with $\alpha=1.0$, Rand Erasing~\cite{zhong2020random} with probability of 25\% and Stochastic Depth with a drop-path rate of 30\%, following the recent works~\cite{swin,deit,bao2021beit,liu2022convnet}. The RepLKNet-31B reported in Sec.~\ref{sect-imgnet} is trained with the same configurations except the epoch number of 300 and drop-path rate of 50\%.
	
	For finetuning the 224$\times$224-trained RepLKNet-31B with 384$\times$384, we use 32 GPUs, a batch size of 32 per GPU, initial learning rate of $4\times 10^{-4}$, cosine annealing, 1-epoch warm-up, 30 epochs, model EMA (Exponential Moving Average) with momentum of $10^{-4}$, the same RandAugment as above but \emph{no CutMix nor mixup}.

	\subsection*{ImageNet-22K Pretraining and 1K Finetuning}
	
	For pretraining RepLKNet-31B/L on ImageNet-22K, we use 128 GPUs and a batch size of 32 per GPU to train for 90 epochs with a drop-path rate of 10\%. The other configurations are the same as the aforementioned ImageNet-1K pretraining. 
	
	Then for finetuning RepLKNet-31B with 224$\times$224, we use 16 GPUs, a batch size of 32 per GPU, drop-path rate of 20\%, initial learning rate of $4\times 10^{-4}$, cosine annealing, model EMA with momentum of $10^{-4}$ to finetune for 30 epochs. Note again that we use the same RandAugment as above but \emph{no CutMix nor mixup}.
	
	For finetuning RepLKNet-31B/L with 384$\times$384, we use 32 GPUs and a batch size of 16 per GPU, and the drop-path rate is raised to 30\%.

	\subsection*{RepLKNet-XL and Semi-supervised Pretraining}
	
	We continue to scale up our architecture and train a ViT-L~\cite{vit} level model named RepLKNet-XL. We use $B=[2,2,18,2]$, $C=[256,512,1024,2048]$, $K=[27,27,27,13]$, and introduce inverted bottleneck with expansion ratio of 1.5 to each RepLK Block.
	During pretraining, we use a private semi-supervised dataset named \emph{MegData73M}, which contains 38 million labeled images and 35 million unlabeled ones. Labeled images come from public and private classification datasets such as ImageNet-1K, ImageNet-22K and Places365~\cite{zhou2017places}. Unlabeled images are selected from YFCC100M~\cite{thomee2016yfcc100m}. We design a multi-task label system according to \cite{ghiasi2021multi}, and utilize soft \textit{pseudo} labels which are offline generated by multiple task-specific ViT-Ls wherever human annotations are unavailable. We pretrain our model for up to 15 epochs with similar configurations as ImageNet-1K pretraining. We do \emph{not use CutMix or mixup}, decrease drop-path rate to 20\%, and use a lower initial learning rate of $1.5 \times 10^{-3}$ and a total batch size of 2048. Structural Re-parameterization is omitted because it only brings less than 0.1\% performance gain on such a large-scale dataset. In other words, we observe that the inductive bias (re-parameterization with small kernels) becomes less important as the data become bigger, which is similar to the discoveries reported by ViT~\cite{vit}.
	
	We finetune on ImageNet-1K with input resolution of 320$\times$320 for 30 epochs following BeiT~\cite{bao2021beit}, except for a higher learning rate of $10^{-4}$ and stage-wise learning rate decay of 0.4. Finetuning with a higher resolution of 384$\times$384 brings no further improvements. For downstream tasks, we use the default training setting except for a drop-path rate of 50\% and stage-wise learning rate decay.
	
	\section*{Appendix B: Visualizing the ERF}
	
	Formally, let $\mathrm{I} (n\times 3\times h \times w)$ be the input image, $\mathrm{M} (n\times c\times h^\prime\times w^\prime)$ be the final output feature map, we desire to measure the contributions of every pixel on $\mathrm{I}$ to the central points of every channel on $\mathrm{M}$, \ie, $\mathrm{M}_{:,:,h^\prime/2, w^\prime/2}$, which can be simply implemented via taking the derivatives of $\mathrm{M}_{:,:,h^\prime/2, w^\prime/2}$ to $\mathrm{I}$ with the auto-grad mechanism. Concretely, we sum up the central points, take the derivatives to the input as the pixel-wise contribution scores and remove the negative parts (denoted by $\mathrm{P}$). Then we aggregate the entries across all the examples and the three input channels, and take the logarithm for better visualization. Formally, the aggregated contribution score matrix $\mathrm{A} (h\times w)$ is given by
	\begin{equation}
	\mathrm{P} = \text{max}(\frac{\partial (\sum_{i}^{n}\sum_{j}^{c}\mathrm{M}_{i,j,h^\prime/2, w^\prime/2})}{\partial \mathrm{I}}, 0) \,, 
	\end{equation}
	\vskip -0.2in
	\begin{equation}
	\mathrm{A} = \log_{10} (\sum_{i}^{n}\sum_{j}^{3}\mathrm{P}_{i,j,:, :} + 1) \,.
	\end{equation}
	
	Then we respectively rescale $\mathrm{A}$ of each model to $[0,1]$ via dividing the maximum entry for the comparability across models.

	\section*{Appendix C: Dense Convolutions \vs Dilated Convolutions}
	\label{sec:comp_dilated}
	
	As another alternative to implement large convolutions, \emph{dilated convolution} \cite{yu2015multi,chen2017deeplab} is a common component to increase the \emph{receptive field (RF)}. However, Table~\ref{table-mob2-dilation} shows though a depth-wise dilated convolution may have the same maximum RF as a depth-wise dense convolution, its representational capacity is much lower, which is expected because it is mathematically equivalent to a \emph{sparse} large convolution. Literature (\eg, \cite{wang2018understanding,xie2021segformer}) further suggests that dilated convolutions may suffer from \emph{gridding problem}. We reckon the drawbacks of dilated convolutions could be overcome by mixture of convolutions with different dilations, which will be investigated in the future. 
	
	\begin{table}
		\caption{MobileNet V2 with all regular DW 3$\times$3 layers replaced by 3$\times$3 dilated layers.}
		\label{table-mob2-dilation}
		\vspace{-0.2in}
		\begin{center}
			\small
			\begin{tabular}{lcccccc}
				\hline
				Max RF &	Kernel size		& Dilation			&	\makecell{ImageNet acc}     &   Params   & FLOPs	\\
				\hline
				9					&	9$\times$9		&	-				&	72.67	    & 4.0M  &   319M  \\
				9					&	3$\times$3		&	4				&	57.23	    & 3.5M  &   300M  \\
				\hline
				13					&	13$\times$13	&	-				&	72.53	    & 4.6M  &   361M  \\
				13					&	3$\times$3		&	6				&	51.21	    & 3.5M  &   300M  \\
				\hline
			\end{tabular}
		\end{center}
		\vspace{-0.2in}
	\end{table}

	\section*{Appendix D: Visualizing the Kernel Weights with Small-Kernel Re-parameterization}
	
	\begin{figure}[t]
		\centering
		\begin{subfigure}{0.66\linewidth}
			\includegraphics[width=\linewidth]{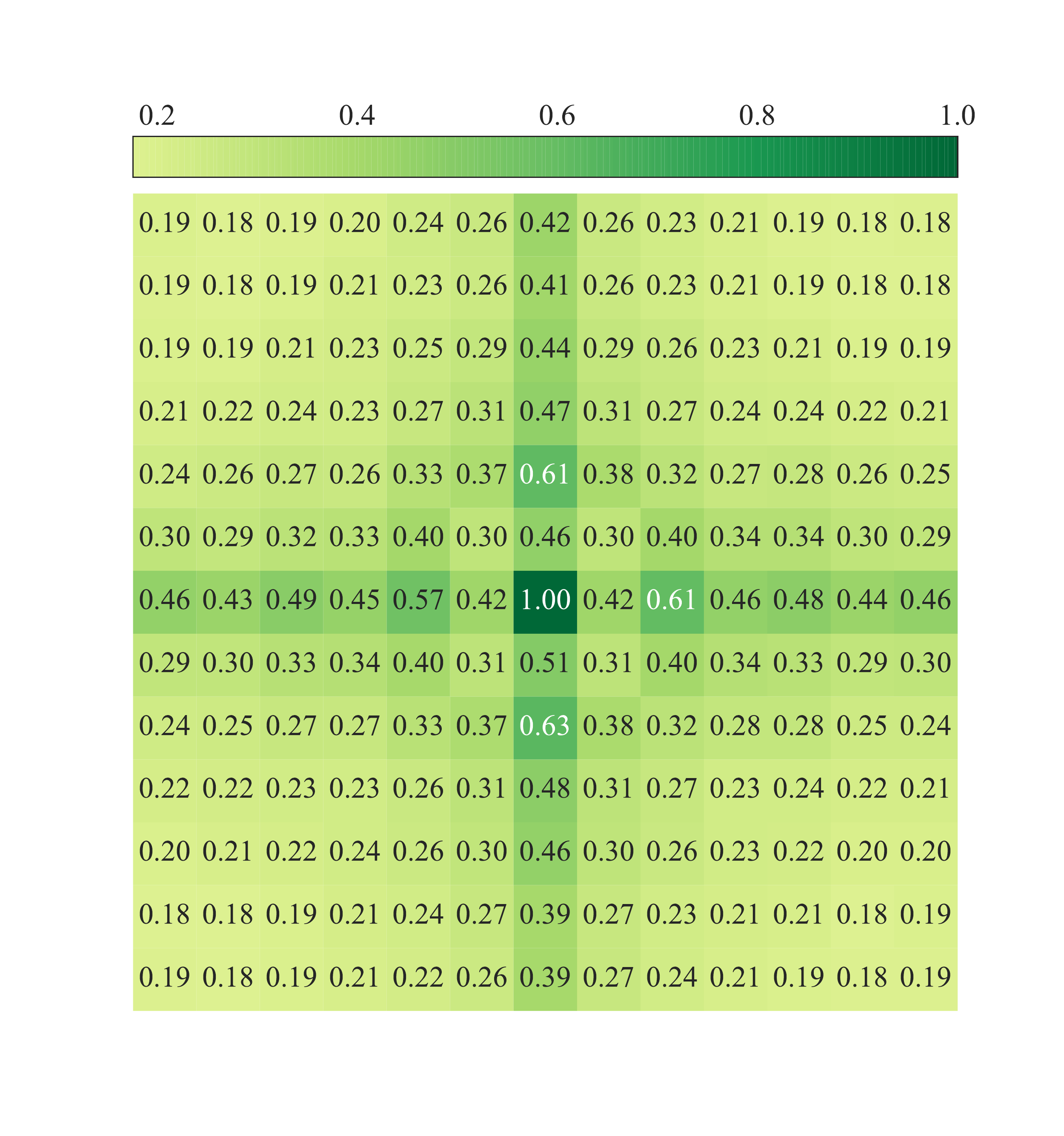}
			\label{fig-kernel-before} 
			\vskip -0.3in
			\caption{Before re-param.}
		\end{subfigure}
		\begin{subfigure}{0.66\linewidth}
			\includegraphics[width=\linewidth]{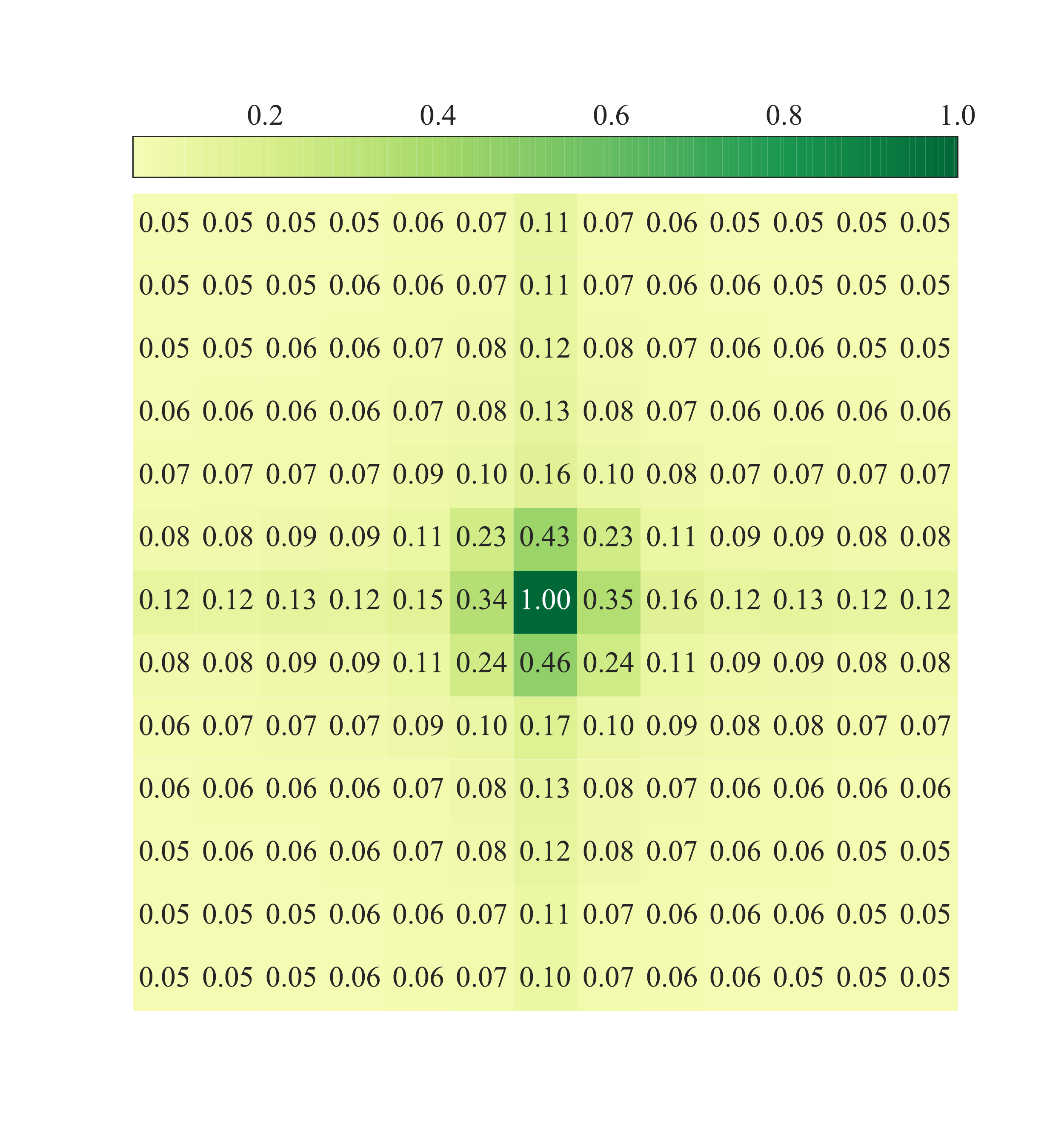}
			\label{fig-kernel-after}
			\vskip -0.3in
			\caption{After re-param.}
		\end{subfigure}
		\begin{subfigure}{0.66\linewidth}
			\includegraphics[width=\linewidth]{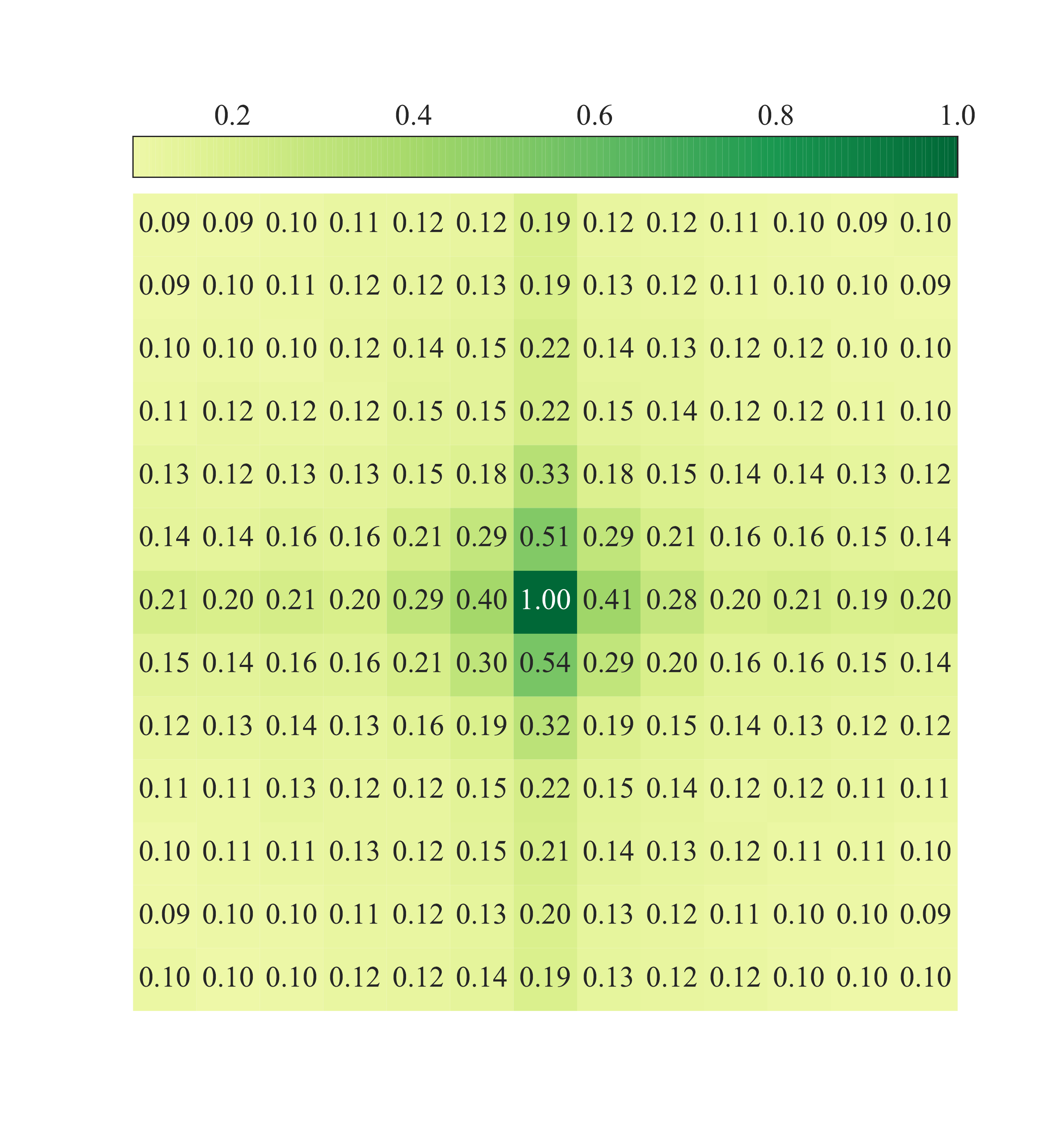}
			\label{fig-kernel-noreparam}
			\vskip -0.3in
			\caption{Without re-param.}
		\end{subfigure}
		\caption{Parameters of 13$\times$13 kernels in MobileNet V2 aggregated into 13$\times$13 matrices.}
		\label{fig-tradeoff}
		\vskip -0.15in
	\end{figure}
	
	We visualize the weights of the re-parameterized 13$\times$13 kernels. Specifically, we investigate into the MobileNet V2 models both with and without 3$\times$3 re-parameterization. As Shown in Sec.~\ref{sect-3} (Guideline 3) , the ImageNet scores are 73.24\% and 72.53\%, respectively. We use the first stride-1 13$\times$13 conv in the last stage (\ie, the stage with input resolution of 7$\times$7) as the representative, and aggregate (take the absolute value and sum up across channels) the resultant kernel into a 13$\times$13 matrix, and respectively rescale to $[0, 1]$ for the comparability. For the model with 3$\times$3 re-param, we show both the original 13$\times$13 kernel (only after BN fusion) and the result after re-param (\ie, adding the 3$\times$3 kernel onto the central part of 13$\times$13). For the model without re-param, we also fuse the BN for the fair comparison.
	
	We observe that every aggregated kernel shows a similar pattern: the central point has the largest magnitude; generally, points closer to the center have larger values; and the ``skeleton'' parameters (the 13$\times$1 and 1$\times$13 criss-cross parts) are relatively larger, which is consistent with the discovery reported by \emph{ACNet}~\cite{ding2019acnet}. But the kernel with 3$\times$3 re-param differs in that the central 3$\times$3 part of the resultant kernel is further enhanced, which is found to improve the performance.

\end{document}